\pdfoutput=1

\documentclass[11pt]{article}

\usepackage[preprint]{acl}

\usepackage{times}
\usepackage{latexsym}

\usepackage[T1]{fontenc}

\usepackage[utf8]{inputenc}

\usepackage{microtype}

\usepackage{inconsolata}

\usepackage{graphicx}
\usepackage{subcaption}

\usepackage{booktabs}
\usepackage{multirow}
\usepackage{multicol}
\usepackage{CJKutf8}
\usepackage{mathtools}
\usepackage{amsmath}
\usepackage{makecell}
\usepackage{fontawesome}

\newcommand\ch[1]{\begin{CJK}{UTF8}{gbsn}#1\end{CJK}}

\usepackage{xcolor}
\definecolor{neu}{rgb}{0.45, 0.45, 0.45} 
\definecolor{orp}{rgb}{0.45, 0, 0.70} 
\definecolor{decrease}{rgb}{0.85, 0, 0} 
\definecolor{increase}{rgb}{0, 0.7, 0} 
\definecolor{neg}{rgb}{0.85, 0, 0} 
\definecolor{pos}{rgb}{0, 0.7, 0} 

\newcommand\eg{\textit{e.g.,}}

\newcommand{\beq}{\begin{equation}}
\newcommand{\eeq}{\end{equation}}
\newcommand{\beqnn}{\begin{equation*}}
\newcommand{\eeqnn}{\end{equation*}}
\newcommand{\beqy}{\begin{eqnarray}}
\newcommand{\eeqy}{\end{eqnarray}}
\newcommand{\beqynn}{\begin{eqnarray*}}
\newcommand{\eeqynn}{\end{eqnarray*}}
\newcommand{\bit}{\begin{itemize}}
\newcommand{\eit}{\end{itemize}}
\newcommand{\ben}{\begin{enumerate}}
\newcommand{\een}{\end{enumerate}}
\newcommand{\bex}{\begin{example}}
\newcommand{\eex}{\end{example}}

\newcommand{\balg}[1]{\begin{algorithm} \caption{#1}}
\newcommand{\ealg}{\end{algorithm}}

\newcommand{\balgc}{\begin{algorithmic}[1]}
\newcommand{\ealgc}{\end{algorithmic}}

\newcommand{\bary}{\begin{array}}
\newcommand{\eary}{\end{array}}
\newcommand{\bmx}{\begin{bmatrix}}
\newcommand{\emx}{\end{bmatrix}}
\newcommand{\bsmx}{\left[\begin{smallmatrix}}
\newcommand{\esmx}{\end{smallmatrix}\right]}
\newcommand{\bmxc}[1]{\left[\begin{array}{@{}#1@{}}}
\newcommand{\emxc}{\end{array}\right]}
\newcommand{\bcn}{\begin{center}}
\newcommand{\ecn}{\end{center}}

\title{\raisebox{-2mm}{\includegraphics[width=0.7cm]{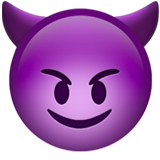}}
FanChuan: A Multilingual and Graph-Structured Benchmark For Parody Detection and Analysis}

\author{
Yilun Zheng$^1$,
Sha Li$^1$,
Fangkun Wu$^1$,
Yang Ziyi$^1$,
Lin Hongchao$^1$,
Zhichao Hu$^1$,
Cai Xinjun$^1$,\\
\textbf{Ziming Wang}$^1$,
\textbf{Jinxuan Chen}$^1$,
\textbf{Sitao Luan}$^{2*}$,
\textbf{Jiahao Xu}$^{1*}$,
\textbf{Lihui Chen}$^1$\thanks{
Corresponding author:
sitao.luan@mail.mcgill.ca,
jiahao004@e.ntu.edu.sg,
elhchen@ntu.edu.sg. 
}
\\
$^1$Nanyang Technological University, Centre for Info. Sciences and Systems,\\
$^2$Mila - Quebec Artificial Intelligence Institute.
}

\begin{document}
\maketitle

\begin{abstract}

Parody is an emerging phenomenon on social media, where individuals imitate a role or position opposite to their own, often for humor, provocation, or controversy. Detecting and analyzing parody can be challenging and is often reliant on context, yet it plays a crucial role in understanding cultural values, promoting subcultures, and enhancing self-expression. However, the study of parody is hindered by limited available data and deficient diversity in current datasets. To bridge this gap, we built seven parody datasets from both English and Chinese corpora, with 14,755 annotated users and 21,210 annotated comments in total. To provide sufficient context information, we also collect replies and construct user-interaction graphs to provide richer contextual information, which is lacking in existing datasets. With these datasets, we test traditional methods and Large Language Models (LLMs) on three key tasks: (1) parody detection, (2) comment sentiment analysis with parody, and (3) user sentiment analysis with parody. Our extensive experiments reveal that parody-related tasks still remain challenging for all models, and contextual information plays a critical role. Interestingly, we find that, in certain scenarios, traditional sentence embedding methods combined with simple classifiers can outperform advanced LLMs, \eg{} DeepSeek-R1 and GPT-o3, highlighting parody as a significant challenge for LLMs.
Our code and data is available at \url{https://github.com/Lisaaa1017/Fanchuan}.
\end{abstract}
\section{Introduction}

\begin{figure}[t]
  \includegraphics[width=\columnwidth]{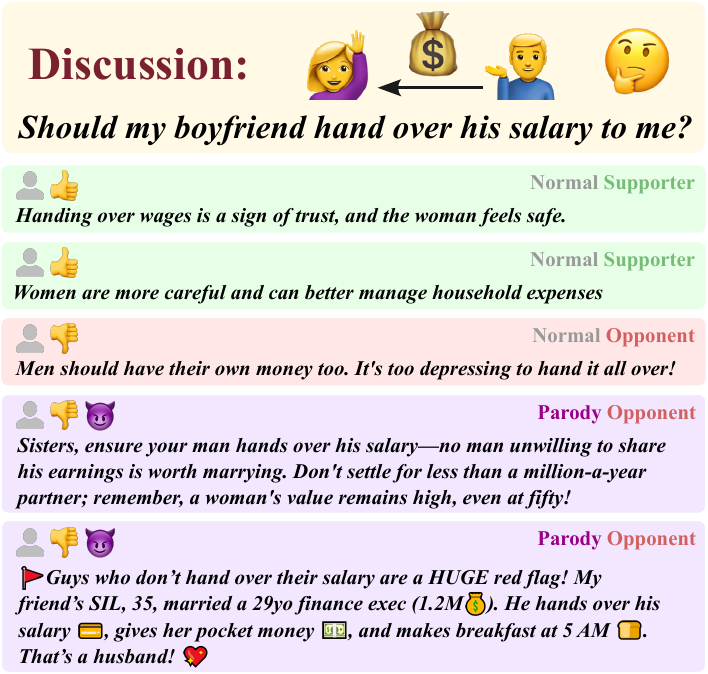}
  \caption{People debate online about the topic, ``\textit{Should my boyfriend hand over his salary to me?}'' Some users explicitly support or oppose this viewpoint, while others implicitly express their stance through parody, using humor or even subtle blackmail to make their point.}\vspace{-0.48cm}
  \label{fig:example_parody}
\end{figure}

Parody in social media\footnote{Also known as "\ch{反串}" or "FanChuan" in Chinese online social media.} is a form of humor or satire, which uses exaggerated or absurd imitations for critique or entertainment \citep{parody_definition}. It has become popular around some controversial topics in recent years, especially among the young generation \citep{parody_young,parody_young2}. For example, as shown in Figure \ref{fig:example_parody}, the question ``\textit{Should my boyfriend hand over his salary to me?}'' has sparked intense debate \citep{news}. While some users clearly express their views as \textcolor{neu}{neutral}, \textcolor{pos}{supportive}, or \textcolor{neg}{opposed}, others adopt a \textcolor{orp}{parody} tone, mockingly pretending to support the stance with exaggerated statements like, ``\textit{Guys who don’t hand over their salary are a HUGE red flag...}'', which subtly opposes it. This tactic can attract attention and provoke reactions through humor, making people reflect their opinions. Similar to irony or sarcasm \citep{topic_gender_EPIC}, parody also expresses the opinion opposite to its appearance. However, it emphasizes playful, entertaining, and exaggerated mimicry of a character, making the underlying critique more accessible and engaging to the audience.

The real meaning behind parody is highly culture-dependent. Therefore, the analysis of parody can offer unique insights in understanding the corresponding cultural values. The spread of parody on internet also fosters a diverse linguistic culture \citep{menghini2024digital}. People can share their distinct views on society, political, or cultural topics in a humorous and engaging manner, encouraging global and cross-cultural dialogue. In addition, parody plays a crucial role in the formation of subcultures \citep{willett2009parodic,booth2014slash}. Parody comments not only create distinct communities, but also mirror the values and identities of online users. For younger generations, parody comments have become a way of self-expression, which help to define their uniqueness, build connection with others, and form social circles. Gradually, it has become a shared language and a set of symbols for the growth of internet subcultures. 

Despite the widespread popularity of parody, there is a lack of high-quality datasets that capture parody comments with different topics and languages \citep{parody_dataset}, restricting the more general and inclusive analysis in various contexts. To fill this gap, we propose FanChuan, a parody benchmark with high quality in three key aspects: \textbf{high diversity, rich contexts, and precise annotations}. \textbf{First}, we enhance diversity by collecting data from multiple sources (both Chinese and English corpora), a wide range of topics, and various social media platforms. Such broad coverage allows us to conduct more sufficient, balanced and fair evaluations of models. \textbf{Second}, we construct richer context information by building the relationship between comments and their replies as heterogeneous graphs. Unlike previous studies that only focus on textual \citep{text_zhang} or dialogue \citep{dialogue_bamman,dialogue_wang} content, the graph-structured context enables the exploitation of relational information, which is found to be fairly valuable later. \textbf{Third}, since parody labeling is quite challenging and disagreements among annotators can easily arise, we ensure the quality of annotation by employing native speakers to label the parody and sentiment of each comment. Additionally, we have expert judges to resolve any disagreement and Large Language Models (LLMs) to refine the annotation results, ensuring consistency and reliability. As a result, we have created \textbf{seven} datasets, with \textbf{14,755} annotated users and \textbf{21,210} annotated comments in total, enabling comprehensive experiments and analyses. 


With the new datasets, we evaluate embedding-based methods \citep{RoBERTa}, incongruity-based methods \citep{SarcPrompt}, outlier detection methods \citep{IsolationForest}, graph-based methods \citep{GCN}, and Large Language Models (LLMs) \citep{GPT4} on FanChuan with three parody related tasks: parody detection, comment sentiment classification with parody, and user sentiment classification with parody. Our results indicate that \textbf{(1)} parody-related tasks are challenging for all models, and even LLMs fail to consistently outperform traditional embedding-based approaches; \textbf{(2)} model performance of sentiment classification drops significantly on comments exhibiting parody behavior compared to those without parody; \textbf{(3)} incorporating commented objects as contextual information greatly enhances parody detection performance; \textbf{(4)} reasoning LLMs fail to outperform non-reasoning LLMs on parody detection. To our best knowledge, the existing studies on parody\citep{parody_dataset,willett2009parodic} are all from pre-LLMs era, and we are the first to evaluate the performance of LLMs on parody detection. In summary, our contributions are summarized as follows:

\begin{itemize}
    \item We introduce FanChuan, a parody benchmark that includes seven datasets from both Chinese and English corpora, containing 21,210 annotated comments and 14,755 annotated users.
    \item We leverage heterogeneous graphs to model user interaction relationships, providing richer contextual information compared to previous datasets.
    \item We comprehensively evaluate five types of methods, including embedding-based methods, inconsistency-based methods, outlier detection methods, graph-based methods, and LLMs, on three parody-related tasks.
    \item Our findings reveal that parody-related tasks are challenging and LLMs cannot always outperform traditional embedding-based methods. Additionally, we show that reasoning LLMs generally underperform non-reasoning LLMs in parody detection.
\end{itemize}

\section{FanChuan}

\begin{figure*}[htbp]
\centering
  \includegraphics[width=1\linewidth]{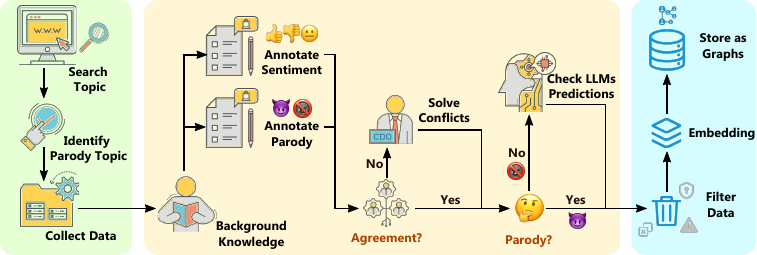}
  \caption{The pipeline for the construction of FanChuan, which includes three key steps: data collection (left), annotation (middle), and preprocessing (right).}\vspace{-0.48cm}
  \label{fig:data_construction}
\end{figure*}

In this section, we will introduce the details about FanChuan. Specifically, in Section \ref{sec:datast_construction}, we introduce the dataset construction process, including data collection, annotation and preprocessing. These steps ensure high diversity, precise annotations, and rich contexts within our dataset. In Section \ref{sec:problem_definition}, we propose three parody-related tasks for model evaluations.

\subsection{Dataset Construction}\label{sec:datast_construction}

As illustrated in Figure \ref{fig:data_construction}, the data construction process for FanChuan involves three steps: data collection, annotation, and preprocessing. Then we introduce the details of each step as follows.

\paragraph{Data collection} To ensure a comprehensive evaluation, we ensure \textbf{high diversity} in our benchmark by selecting a wide range of topics from both Chinese and English corpora. Given that parody often emerges around controversial issues, we begin by focusing on topics or recent events that have sparked intense debates on social media. To select the post that includes adequate parody comments, we randomly sample a subset of its comments to determine the proportion of parody content. If more than $3\%$ of the comments are identified as parody, we classify it as suitable for further collection. To capture the most relevant content, we use keyword search to identify prominent posts, then collect their comments, replies, and associated content.

\paragraph{Data Annotation} Labeling parody presents a significant challenge, not only because it requires a high familiarity with the content and culture \citep{banziger2005role}, but also due to potential disagreements of understanding among annotators from diverse backgrounds \citep{dress2008regional}. To ensure \textbf{precise annotations} in FanChuan, the annotation process includes five steps: 
\textbf{(1)} To provide accurate and culturally relevant insights, we assign native speakers to annotate Chinese and English datasets, respectively. Annotators are then asked to review relevant materials to enhance their understanding before starting the annotation process.
\textbf{(2) Sentiment Annotation.} Annotators classify the sentiment of a given comment or user by answering the question: \textit{``Does this comment or user support, oppose, or remain neutral regarding to this statement?''}
\textbf{(3) Parody Annotation.} After sentiment classification, annotators are asked to determine whether a comment is a parody by answering the question: \textit{``Is this comment a parody or not?''} During both sentiment and parody annotation stages, annotators are provided with relevant comments and context to ensure accurate labeling.
\textbf{(4) Resolving Discrepancies.} Each comment receives a final label based on the majority vote of three annotators. If consensus is not reached, the most knowledgeable annotator on the relevant topic or event reassesses the labels.
\textbf{(5) Verification.} To minimize errors in parody annotations, an experienced annotator reviews all comments labeled as parody. Note that this annotator will also double-check the comments that are labeled as parody by LLMs but not labeled by human annotators. 

\paragraph{Data preprocessing} To ensure data quality, we first delete any content or comments that contain irrelevant, sensitive, personal, or hazardous information. We provide three types of embeddings: Bag of Words (BoW) \citep{BoW}, Skip-gram \citep{Skip-gram}, and RoBERTa \citep{RoBERTa}. Given that the context of parody forms a network structure, we store the data as heterogeneous graphs as shown in Figure \ref{fig:ORP_graph}, where the nodes represent users and posts, and there are two types of edges to represent two types of relations: user-comments-post, and user-comments-user. Compared with existing datasets \citep{dialogue_bamman, ptaek2014sarcasm} that focus solely on content or dialogue, such graph-structured data enables deeper understanding of parody with \textbf{richer contexts}, including 2-hop neighbors and higher-order relationships.

Finally, as shown in Table \ref{tab:dataset_statistics}, we constructed seven datasets from both Chinese and English corpora, encompassing multiple topics, with a total of 14,755 annotated users and 21,210 annotated comments. Our analysis reveals that parody comments constitute only a small proportion of the total comments across all datasets. For detailed description and background information of each dataset, please refer to Appendix \ref{apd:dataset_details}.

\begin{table*}[htbp]
\resizebox{1\hsize}{!}{
\begin{tabular}{ccc|ccc|cc}
\toprule
\multirow{2}{*}{\textbf{Dataset}} & \multirow{2}{*}{\textbf{Topic}} & \multirow{2}{*}{\textbf{Language}} & \multicolumn{3}{c|}{\textbf{Comment}} & \multicolumn{2}{c}{\textbf{User}} \\ \cline{4-8}
& & & \textbf{\#Num} & \textbf{\#Parody/\#Normal} & \textbf{\#Pos / \#Neg / \#Neu} & \textbf{\#Num} & \textbf{\#Pos / \#Neg / \#Neu} \\
\midrule
Alibaba-Math & Education & Chinese & 8353 & 489 / 7864 & 1831 / 1509 / 5013 & 5247 & 1397 / 1044 / 2806  \\
BridePrice & Social & Chinese & 1774 & 84 / 1690 & \ \ 20 / 385 / 1369 & 1254 & 17 / 341 / 896 \\
DrinkWater & Technology & Chinese & 3659 & 113 / 3546 & 378 / 384 / 2897 & 3204 & 349 / 353 / 2502 \\
CS2 & Game & Chinese & 3196 & 196 / 3000 & 169/480/517/25/2005* & 2093 & \text{117/372/385/19/1200}*\\
CampusLife & Life & English & 1206 & 89 / 1117 & 41 / 201 / 964 & 569 & 30 / 131 / 408 \\
Tiktok-Trump & Politics & English & 1634 & 97 / 1537 & 150 / 495 / 989 & 1237 & 127 / 434 / 676\\
Reddit-Trump & Politics & English & 1388 & 171 / 1217 & 169 / 678 / 541 & 1151 & 149 / 594 / 408 \\
\bottomrule
\end{tabular}
}
\caption{Dataset Statistics. *In particular, for CS2, there are five types of sentiment labels: support for G2 (a gaming club), support for NAVI (another gaming club), opposition to G2, opposition to NAVI, and neutral.}\vspace{-0.4cm}
\label{tab:dataset_statistics}
\end{table*}


\subsection{Problem Definition}\label{sec:problem_definition}

\begin{figure*}[htbp]
\centering
  \includegraphics[width=1\linewidth]{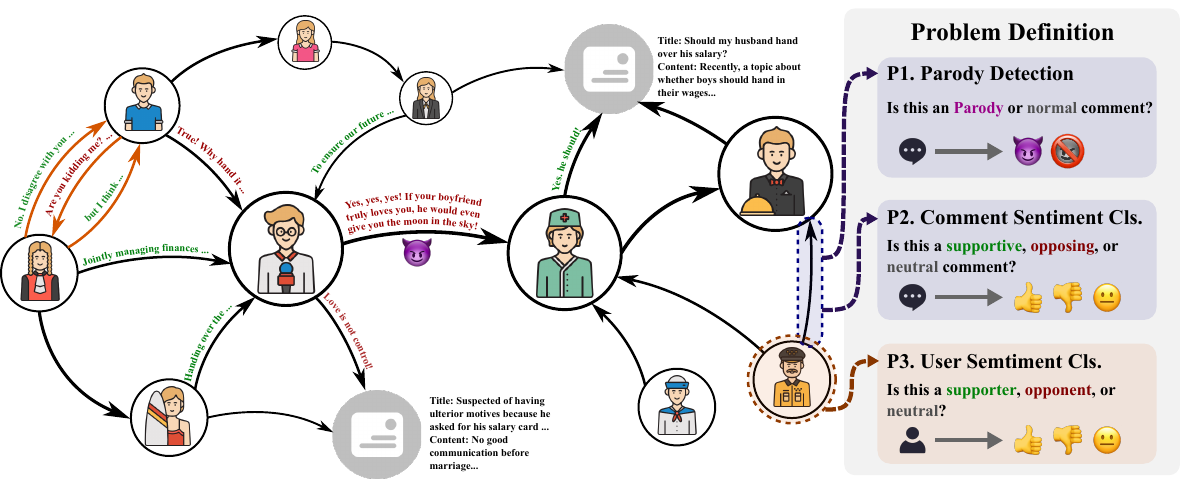}
  \caption{Examples of a parody dataset as a heterogeneous graph.}\vspace{-0.4cm}
  \label{fig:ORP_graph}
\end{figure*}

As shown in Figure \ref{fig:ORP_graph}, we utilize Heterogeneous Information Networks (HINs) to structure our datasets, representing the relational information in content and comments. Each HIN comprises two types of nodes: user nodes and post nodes, along with two types of edges: user comments to posts and user comments to users\footnote{A comment on another comment inherently forms an edge linking to another edge, which cannot be directly represented in a graph. Instead, we connect such comments to the target user, as they reflect that user's traits or viewpoints.}. Each edge is directed, with the source being the user and the target either a post or another user. As shown by the \textcolor{orange}{orange} edges on the right in Figure \ref{fig:ORP_graph}, multiple edges may exist between two nodes due to several rounds of replies among these users. This results in a directed multigraph \citep{gross2003handbook}. Each edge or node is associated with text as features. We then introduce three tasks as follows.

\paragraph{P1. Parody Detection} Parody detection aims to identify whether a comment is \textcolor{orp}{parody} or \textcolor{neu}{normal}. In HINs, this can be framed as a binary classification task on edges. Given that parody comments represent a small fraction of all comments, this task can also be considered as outlier detection.

\paragraph{P2. Comment Sentiment Classification} Like parody detection, comment sentiment classification aims to categorize comments into three sentiment labels: \textcolor{pos}{positive}, \textcolor{neg}{negative}, and \textcolor{neu}{neutral}.

\paragraph{P3. User Sentiment Classification} This task focuses on classifying users' sentiment as either a \textcolor{pos}{supporter}, \textcolor{neg}{opponent}, or \textcolor{neu}{neutral}. Unlike the edge classification tasks discussed earlier, this is a node classification task in HINs.

\paragraph{Remarks} We introduce sentiment classification tasks due to the complexity of the scenarios that include parody comments \cite{bull2010automatic}. In the context of parody, these tasks serve as a comprehensive measure to assess the effectiveness of current models in handling parody-related tasks, which will be introduced in the next section.


\section{Experiments}
\label{sec:Experiments}


\subsection{Settings}\label{sec:exp_settings}

We split all the comment data into training, validation, and test sets with a ratio of 40\%/30\%/30\%. We consider parody detection as a binary classification problem use F1 score for the evaluation. We model the comment and user sentiment classification with parody as multi-class classification problems, and use Macro-F1 to measure the model performance.
For comprehensive evaluation and analysis, we test five types of approach in our experiments:

\textbf{(1) Embedding-based methods.} This category includes Bag-of-Words (BoW) \citep{BoW}, Skip-gram \citep{Skip-gram}, and RoBERTa \citep{RoBERTa}, all of which utilize Multi-Layer Perceptron (MLP) classifiers. These methods are widely used and can provide general text representations to capture linguistic patterns and semantics.

\textbf{(2) Inconsistency-based methods.} These methods are commonly used for irony detection and we assess BNS-Net \citep{BNS-Net}, DC-Net \citep{DC-Net}, QUIET \citep{QUIET}, and SarcPrompt \citep{SarcPrompt}. Similar to irony or sarcasm, parody usually contains inconsistencies between literal and intended meaning, and thus, the evaluation of these methods are necessary.

\textbf{(3) Outlier detection methods.} This category includes Isolation Forest \citep{IsolationForest}, the Z-Score Method \citep{Z-Score}, and One-Class SVM \citep{OneclassSVM}. Similar to outlier detection tasks, where data is highly imbalanced, parody only accounts for around 5\%-10\% of all comments and tremendously deviates from the normal comment patterns, which makes outlier detection methods quite relevant.

\textbf{(4) Graph-based methods.} Since (graph-structured) context information is highly important for parody understanding, and to capture complex structural information in user interaction graphs, Graph Neural Networks (GNNs) could be used for user sentiment classification. Three types of classical GNNs are used: Graph Convolutional Networks (GCN) \citep{GCN}, Graph Attention Networks (GAT) \citep{GAT}, and GraphSAGE \citep{GraphSAGE}.

\textbf{(5) Large Language Models (LLMs).} We evaluate models such as ChatGPT-4o (and 4o-mini) \citep{GPT4} from OpenAI, Claude 3.5 \citep{Claude} from Anthropic, Qwen 2.5 \citep{Qwen2.5} from Alibaba, and DeepSeek-V3 \citep{DeepSeek} from DeepSeek. Given the strong reasoning capabilities and contextual understanding of LLMs in NLP-related tasks, we assess their performance in parody detection under a zero-shot setting.

\subsection{Performance Comparison}\label{sec:exp_performance}
The evaluation results on the three parody-related tasks are shown in Table \ref{tab:ORP_detection_new}, \ref{tab:sentiment_detection_new},  \ref{tab:user_sentiment_detection_new}. The best and runner-up methods for each dataset are highlighted in \textbf{bold} and \underline{underlined}, respectively. Then, the detailed comparison and analysis are as follows.

\begin{table*}[h]
\centering
\resizebox{1\textwidth}{!}{
\begin{tabular}{ll|ccccccc|c}
\toprule
\textbf{Paradigm} & \textbf{Method} & \textbf{Alibaba.} & \textbf{Bride.} & \textbf{Drink.} & \textbf{CS2} & \textbf{Campus.} & \textbf{Tiktok.} & \textbf{Reddit.} & \textbf{Ave. Rank} \\
\midrule
\multirow{3}{*}{\makecell[l]{Embedding\\-based}} 
& BoW+MLP & 10.17 & 15.83 & 9.06 & 15.93 & 11.20 & 13.71 & 16.91 & 9.57  \\
& Skip-gram+MLP & 14.16 & 17.50 & 14.55 & 17.29 & 10.40 & 15.43 & 14.85 & 7.86  \\
& RoBERTa+MLP & 14.30 & 19.17 & 13.33 & 16.61 & 16.52 & 12.00 & 23.09 & 6.71  \\
\midrule
\multirow{4}{*}{\makecell[l]{Inconsistency\\-based}} 
& BNS-Net & 13.62 & 12.31 & \underline{16.67} & 20.00 & 28.17 & 24.86 & 16.67 & 6.57  \\
& DC-Net & 13.54 & 10.53 & \textbf{17.39} & 14.37 & 14.04 & 9.38 & 24.16 & 9.29  \\
& QUIET & \underline{15.98} & 10.75 & 4.94 & 7.75 & 13.07 & 10.11 & 16.34 & 10.57  \\
& SarcPrompt & 14.20 & \textbf{22.22} & 5.26 & \textbf{21.39} & 26.67 & 15.38 & 15.09 & 6.72  \\
\midrule
\multirow{3}{*}{\makecell[l]{Outlier\\ Detection}} 
& Isolation Forest & 5.93 & 1.18 & 0.90 & 7.14 & 5.62 & 6.15 & 11.70 & 14.71  \\
& RoBERTa+Z-Score & 13.06 & \underline{20.83} & 12.31 & 18.64 & 17.78 & 14.29 & 22.68 & 7.14  \\
& One-Class SVM & 5.81 & 4.71 & 1.79 & 5.64 & 7.82 & 9.14 & 14.99 & 14.14  \\
\midrule
\multirow{5}{*}{LLMs} 
& ChatGPT4o & 15.90 & 13.54 & 8.94 & 18.86 & 34.29 & \textbf{39.51} & \textbf{37.26} & \textbf{3.86}  \\
& ChatGPT4o-mini & 13.73 & 11.06 & 8.91 & 16.00 & \underline{40.00} & \underline{36.41} & \underline{36.90} & 6.14  \\
& Claude3.5 & 13.21 & 12.49 & 8.56 & 16.00 & \textbf{41.24} & 29.96 & 36.45 & 6.71  \\
& Qwen2.5 & 14.88 & 12.44 & 7.81 & 19.38 & 28.89 & 27.70 & 33.29 & 6.14  \\
& DeepSeek-V3 & \textbf{16.17} & 13.24 & 9.19 & \underline{20.45} & 32.55 & 31.10 & 34.34 & \textbf{3.86}  \\
\bottomrule
\end{tabular}
}
\caption{Comparison of model performance in parody detection using F1 score (\%).}
\label{tab:ORP_detection_new}
\end{table*}
\paragraph{Parody Detection.} The results in Table \ref{tab:ORP_detection_new} indicate that: (1) Parody detection is challenging for all models, with most achieving only $10\%\sim40\%$ F1 scores. Even the best-performing methods for \textit{Alibaba.} and \textit{Drink.} reach only $16.17\%$ and $17.39\%$, respectively, highlighting the difficulty of the task. (2) LLMs generally rank higher but struggle with Chinese datasets. Specifically, both of ChatGPT-4o and Deepseek-V3 achieve $3.86$ average rank across all datasets, outperforming other methods. However, traditional methods perform better on Chinese datasets. For instance, SarcPrompt achieves an F1 score of $22.22\%$ on \textit{Bride.} and $21.39\%$ on \textit{CS2}, outperforming the best LLM by a large margin. In addition to the performance comparison, we conduct a case study to further investigate how well LLMs understand parody detection in Appendix \ref{apd:sec_case_study}.

\begin{table*}[h]
\centering
\resizebox{1\textwidth}{!}{
\begin{tabular}{ll|ccccccc|c}
\toprule
\textbf{Paradigm} & \textbf{Method} & \textbf{Alibaba.} & \textbf{Bride.} & \textbf{Drink.} & \textbf{CS2} & \textbf{Campus.} & \textbf{Tiktok.} & \textbf{Reddit.} & \textbf{Ave. Rank} \\
\midrule
\multirow{3}{*}{\makecell[l]{Embedding\\-based}} 
& BoW+MLP & 35.30 & \textbf{40.43} & \underline{48.78} & 27.56 & 32.35 & 33.74 & 37.13 & 8.14  \\
& Skip-gram+MLP & 39.62 & \underline{39.50} & 47.46 & 31.09 & 30.80 & 35.42 & 37.71 & 6.29  \\
& RoBERTa+MLP & 36.91 & 34.48 & 44.17 & 26.02 & \underline{38.87} & 47.56 & 51.66 & 8.00  \\
\midrule
\multirow{4}{*}{\makecell[l]{Inconsistency\\-based}} 
& BNS-Net & 35.48 & 29.40 & 45.66 & 21.13 & 29.71 & 26.47 & 22.08 & 7.29  \\
& DC-Net & 16.07 & 28.87 & 48.66 & 18.89 & \textbf{38.90} & 45.21 & 37.18 & 7.29  \\
& QUIET & 24.34 & 30.26 & 35.52 & 17.65 & 30.05 & 29.51 & 23.95 & 7.00  \\
& SarcPrompt & 28.77 & 28.85 & 33.91 & 19.18 & 35.21 & 40.06 & 22.69 & 5.43  \\
\midrule
\multirow{5}{*}{LLMs} 
& ChatGPT4o & 40.00 & 32.28 & 47.75 & \textbf{37.82} & 32.10 & 51.02 & 51.89 & \underline{4.86}  \\
& ChatGPT4o-mini & \underline{40.01} & 34.27 & \textbf{49.95} & 34.33 & 33.19 & \underline{51.56} & \underline{52.42} & \textbf{4.29}  \\
& Claude3.5 & \textbf{40.53} & 29.89 & 42.99 & 30.70 & 28.31 & 46.03 & 51.92 & 5.71  \\
& Qwen2.5 & 38.46 & 31.83 & 46.14 & \underline{34.78} & 28.38 & 47.55 & 51.93 & 6.86  \\
& DeepSeek-V3 & 35.88 & 28.15 & 43.05 & 32.62 & 36.36 & \textbf{56.26} & \textbf{54.83} & 6.86  \\
\bottomrule
\end{tabular}
}
\caption{Comparison of model performance in comment sentiment classification with parody using Macro-F1 score (\%)}\vspace{-0.4cm}
\label{tab:sentiment_detection_new}
\end{table*}
\begin{table*}[h]
\centering
\resizebox{1\textwidth}{!}{
\begin{tabular}{ll|ccccccc|c}
\toprule
\textbf{Paradigm} & \textbf{Method} & \textbf{Alibaba.} & \textbf{Bride.} & \textbf{Drink.} & \textbf{CS2} & \textbf{Campus.} & \textbf{Tiktok.} & \textbf{Reddit.} & \textbf{Ave. Rank} \\
\midrule
\multirow{3}{*}{\makecell[l]{Embedding\\-based}} 
& BoW+MLP & \underline{46.54} & 37.60 & 46.65 & 29.22 & 32.35 & 35.05 & 31.97 & 7.57 \\
& Skip-gram+MLP & \textbf{46.99} & 38.28 & \underline{50.45} & 31.92 & 32.02 & 38.46 & 32.69 & 6.42 \\
& RoBERTa+MLP & 43.11 & 36.94 & 44.20 & 27.09 & 35.49 & \underline{50.82} & \underline{52.79} & \underline{5.00} \\
\midrule
\multirow{3}{*}{\makecell[l]{Inconsistency\\-based}} 
& BNS-Net & 34.32 & 27.21 & 41.91 & 23.38 & 28.67 & 23.61 & 22.98 & 13.00 \\
& DC-Net & 16.51 & 33.56 & 48.65 & 17.17 & 35.60 & 34.62 & 39.54 & 9.57 \\
& SarcPrompt & 27.72 & \underline{38.54} & 29.51 & 15.62 & 31.45 & 24.48 & 39.10 & 11.29 \\
\midrule
\multirow{3}{*}{\makecell[l]{Graph\\-based}} 
& GCN & 37.69 & \textbf{40.00} & 43.67 & 23.64 & 36.45 & 42.94 & 48.06 & 7.00 \\
& GAT & 38.30 & 38.53 & 43.44 & 23.72 & \underline{37.20} & 42.12 & 50.57 & 6.71 \\
& GraphSAGE & 39.92 & 37.63 & 42.79 & 25.98 & 32.94 & 40.66 & 52.08 & 7.71 \\
\midrule
\multirow{5}{*}{LLMs} 
& ChatGPT-4o & 41.71 & 35.02 & \textbf{51.54} & \textbf{39.19} & 35.89 & 45.87 & 49.01 & \textbf{4.14} \\
& ChatGPT-4o-mini & 40.55 & 30.25 & 45.88 & 34.03 & 31.95 & 45.29 & 51.20 & 6.71 \\
& Claude3.5 & 41.47 & 29.96 & 43.78 & 32.81 & 31.07 & 41.85 & 46.92 & 8.57 \\
& Qwen2.5 & 40.89 & 33.08 & 49.52 & \underline{36.51} & 33.34 & 46.18 & 50.13 & 5.29 \\
& DeepSeek-V3 & 40.00 & 26.37 & 41.55 & 33.61 & \textbf{40.49} & \textbf{54.04} & \textbf{53.22} & 6.00 \\
\bottomrule
\end{tabular}
}
\caption{Comparison of model performance in user sentiment classification with parody using Macro-F1 score (\%).}\vspace{-0.4cm}
\label{tab:user_sentiment_detection_new}
\end{table*}
\paragraph{Sentiment Classification.} Tables \ref{tab:sentiment_detection_new} and \ref{tab:user_sentiment_detection_new} present the model performance in comment and user sentiment classification, respectively. Our findings are as follows: (1) Sentiment classification in the context of parody presents significant challenges. The top-performing models across each dataset achieve F1 scores ranging from 40\% to 50\%, which are notably lower than the performance on traditional sentiment classification benchmarks without parody\citep{senti_benchmark1,senti_benchmark2}. (2) Although LLMs show their superiority over other methods in terms of average rank, they still underperform some traditional approaches on certain datasets. For example, although ChatGPT-4o-mini attains the highest average rank of 4.29 in comment sentiment classification, it performs much worse than BoW+MLP on \textit{Bride.} and DC-Net on \textit{Campus.} (3) Graph-based methods demonstrate strong performance on certain datasets. For example, GCN achieves the best results on \textit{Bride.}, suggesting that the relational context information in user-interaction networks is informative and beneficial for some tasks in sentiment classification.

In general, all the parody-related tasks are challenging for current models and no model can take dominant advantage over others cross all datasets. These observations underscore the need for further study and model development on parody-related tasks.

\subsection{Influence of Context on Parody Detection}\label{sec:exp_context_impact}
\begin{table}[h]
\small
\centering
\begin{tabular}{lccc}
\toprule
\textbf{Method} & \textbf{w/o Context} & \textbf{w. Context} & \textbf{$\Delta$} \\ 
\midrule
BoW+MLP & 13.26 & 15.19 & \textcolor{increase}{+1.93} \\
Skip-gram+MLP & 14.88 & 16.19 & \textcolor{increase}{+2.08} \\
RoBERTa+MLP & 16.43 & 21.23 & \textcolor{increase}{+4.80} \\
ChatGPT4o & 24.04 & 28.53 & \textcolor{increase}{+4.49} \\
ChatGPT4o-mini & 23.29 & 23.99 & \textcolor{increase}{+0.70} \\
Claude3.5 & 22.56 & 23.09 & \textcolor{increase}{+0.53} \\
Qwen2.5 & 20.63 & 18.04 & \textcolor{decrease}{-2.59} \\
DeepSeek-V3 & 22.43 & 24.83 & \textcolor{increase}{+2.40} \\
\bottomrule
\end{tabular}
\caption{Impact of context on parody detection using F1 Score (\%) averaged over seven datasets.}\vspace{-0.4cm}
\label{tab:impact_context}
\end{table}

Since parody detection requires a deep understanding of the background information of a topic, intuitively, the context information should have a strong impact on model performance. Therefore, we introduce relevant background details and target comments (when available), and conduct ablation study to investigate its impact on model performance. In Table \ref{tab:impact_context}, we report the average F1 score across seven datasets, both with and without context. Performance improvements and declines are highlighted in \textcolor{increase}{green} and \textcolor{decrease}{red}, respectively.

Overall, most models benefit from contextual information, with ChatGPT-4o improving significantly from $24.04$ to $28.53$ and RoBERTa+MLP increasing from $16.43$ to $21.23$. Our results are consistent with the observations in \citep{dialogue_bamman,dialogue_wang} that context improves model performance on sarcasm and irony detection. However, Qwen2.5 is the only model that performs worse with added context, suggesting potential limitations in how it processes additional information. These results highlight that while context generally enhances parody detection, its effectiveness varies across models. Please refer to Appendix \ref{apd:impact_context} for more details of the impact of context on each dataset.

\subsection{Influence of Parody to Sentiment Classification}\label{sec:exp_parody_senti}
\begin{table}[h]
\small
\centering
\begin{tabular}{lccc}
\toprule
\textbf{Method} & \textbf{Non-Parody} & \textbf{Parody} & \textbf{$\Delta$} \\ 
\midrule
BoW+MLP & 35.71 & 30.21 & \textcolor{decrease}{-5.50} \\
Skip-gram+MLP & 37.17 & 30.08 & \textcolor{decrease}{-7.09} \\
RoBERTa+MLP & 39.65 & 33.15 & \textcolor{decrease}{-6.50} \\
ChatGPT4o & 42.28 & 26.84 & \textcolor{decrease}{-15.44} \\
ChatGPT4o-mini & 42.68 & 27.03 & \textcolor{decrease}{-15.65} \\
Claude3.5 & 38.98 & 24.87 & \textcolor{decrease}{-14.11} \\
Qwen2.5 & 40.15 & 26.87 & \textcolor{decrease}{-13.28} \\
DeepSeek-V3 & 41.17 & 29.29 & \textcolor{decrease}{-11.89} \\
\bottomrule
\end{tabular}
\caption{Impact of parody on comment sentiment classification using Macro F1 Score (\%) averaged over seven datasets.}\vspace{-0.4cm}
\label{tab:impact_parody_senti}
\end{table}

To confirm that parody adds challenges to sentiment classification, we evaluate model performance using Macro F1 score averaged over seven datasets on comment sentiment classification, and compare the results of parody and non-parody comments. As shown in Table \ref{tab:impact_parody_senti}, the average Macro F1 scores decrease by $5\%$ to $15\%$ across all models, indicating that parody significantly increases the difficulty of sentiment classification. Additionally, we observe that while LLMs outperform embedding-based methods on non-parody comments, their performance deteriorates on parody comments, falling a lot behind embedding-based methods. We speculate that this degradation occurs because these topics are relatively new and LLMs have not encountered such data before, whereas the training process in embedding-based methods allows them to better adapt to the updated knowledge. For more details of the impact of context on each dataset, please refer to Appendix \ref{apd:impact_parody}.

\subsection{Reasoning LLMs in Parody Detection}
\label{sec:exp_reasoning_llms}

\begin{figure}[t]
  \centering
  \includegraphics[width=0.9\columnwidth]{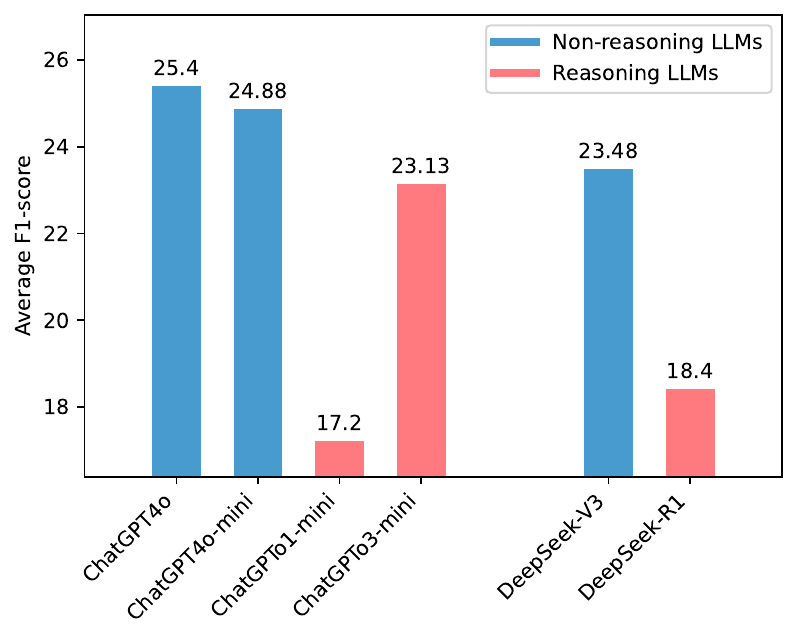}
  \vspace{-0.4cm}
  \caption{Performance comparison between reasoning LLMs and non-reasoning LLMs using average F1 Score (\%) over six datasets.}\vspace{-0.4cm}
  \label{fig:reasoning_llms}
\end{figure}

Recently, there has been a surge in reasoning LLMs \citep{ChatGPT-o1}, which enhance performance by introducing inference-time scaling in the Chain-of-Thought (CoT) \citep{CoT} reasoning process. To assess the impact of reasoning on LLM performance in parody detection, we compared the performance of reasoning LLMs with that of non-reasoning LLMs. Figure \ref{fig:reasoning_llms} presents the average F1 scores of reasoning LLMs, including ChatGPTo1-mini \citep{ChatGPT-o1}, ChatGPTo3-mini \citep{ChatGPT-o3}, and DeepSeek-R1 \citep{DeepSeek-R1}, and non-reasoning LLMs, including ChatGPT4o, ChatGPT4o-mini, and DeepSeek-V3. Surprisingly, unlike math, coding \citep{o1_math} and medical applications \citep{o1_medical}, where reasoning LLMs significantly improve performance, our results show that reasoning LLMs underperform their non-reasoning counterparts. This finding aligns with the conclusion in \citep{yao2024sarcasm}, which suggests that tasks like sarcasm detection do not follow a step-by-step reasoning process. This can explain why CoT does not enhance LLM performance. It indicates that the complexities of parody detection may require alternative strategies beyond reasoning, highlighting the need for further research in this area. Please see Appendix \ref{apd:reasoning_llms} for detailed results on the performance of reasoning LLMs in parody detection.
\section{Related Work}

In this section, we introduce the datasets and detection methods related to parody, as well as its associated topics: sarcasm, irony, and humor.


\subsection{Dataset}\label{sec:rw_dataset}
The datasets for parody and sarcasm cover a diverse array of topics, including politics \citep{topic_politic_Guanchache}, gender \citep{topic_gender_EPIC}, and education \citep{topic_education}. They utilize various modalities, such as text \citep{text_zhang}, speech \citep{speech_ariga}, visual \citep{visual_schif}, and multimodal formats \citep{mm_bedi, mm_maity}. Beyond the content itself, context plays a crucial role in understanding sarcasm or parody \citep{human_wallace}. To enhance contextual information, \citet{dialogue_wang, dialogue_bamman} collect data from dialogues. For annotation, \citet{dialogue_bamman, ptaek2014sarcasm} use user-provided tags as labels, while \citet{riloff2013sarcasm} employ manual annotation. As noted by \citet{survey2024}, the former method requires no human involvement but can lead to noise, as not all users utilize tags. In contrast, the latter approach can yield more generalized labels but may result in significant disagreement among annotators \citep{joshi2016cultural}. In conclusion, most datasets focus on sarcasm detection \citep{topic_politic_Guanchache, text_zhang, mm_maity}, leaving a notable scarcity of parody datasets.

\subsection{Irony or Sarcasm Detection}\label{sec:rw_methods}
Deep learning approaches for detecting parody and sarcasm can be categorized into incongruity-based, sentiment-based, and knowledge-based perspectives \citep{survey2024}. Incongruity-based methods focus on the inherent incongruity that characterizes sarcastic content \citep{riloff2013sarcasm}. For example, \citet{hazarika2018cascade} and \citet{schifanella2016detecting} identify sarcasm by measuring inconsistencies between different targets or modalities. Sentiment-based methods operate on the assumption that there are dependencies between sentiments and sarcasm. \citet{savini2020multi} propose integrating sentiment tasks into the training process alongside sarcasm detection to enhance model performance. To create emotion-rich representations, \citet{babanejad2020affective} incorporate affective and contextual cues. Recognizing that understanding sarcasm can often be implicit, knowledge-based approaches \citep{chen2022commonsense,li2021sarcasm} leverage external knowledge bases. These methods typically involve knowledge extraction, selection, and integration \citep{survey2024}.
\vspace{-0.2cm}
\section{Conclusions}
\vspace{-0.2cm}
In this paper, we introduce FanChuan, a multilingual benchmark for parody detection and analysis, encompassing seven datasets characterized by high diversity, rich contextual information, and precise annotations. Our findings reveal that parody detection remains highly challenging for both LLMs and traditional methods, with particularly poor performance on Chinese datasets. We also observe that contextual information significantly enhances model performance, while parody itself increases the difficulty of sentiment classification. Additionally, our results indicate that reasoning fails to improve LLM performance in parody detection. By filling a critical gap in the study of emerging online phenomena, FanChuan provides valuable insights into cultural values and the role of parody in digital discourse. These findings highlight the limitations of current LLMs, presenting an opportunity for future research to enhance model capabilities in parody detection and analysis.

\clearpage
\section*{Limitations}

While this paper proposes a multilingual parody benchmark and provides an extensive analysis, we acknowledge several limitations that warrant further exploration in future work:

\begin{itemize} 
\item Limited dataset diversity. Although we collect datasets and analyze experimental results in both Chinese and English, the understanding of how parody manifests or how effective current methods are for parody detection in other languages remains unclear. Therefore, further efforts could be made to gather datasets in additional languages to enhance the diversity of parody data.
\item Annotation quality limitations. While we invite multiple annotators and conduct re-checks after labeling, some minor errors may still exist, as annotating parody can be a challenging task. To improve annotation quality in future studies, we will recruit more annotators and provide them with additional background knowledge related to the events before the annotation process. This will help ensure more accurate and consistent annotations.
\item Limited evaluation of Large Language Models (LLMs). In this study, we only test the performance of LLMs on parody-related tasks through prompt-based methods, without fine-tuning. This approach may not fully capture the potential of LLMs. Additionally, only 6 LLMs were evaluated, which is a relatively small number considering the rapid development of these models. Future work should include a broader range of LLMs and explore fine-tuning approaches to better assess their capabilities in parody detection tasks.
\item Limited exploration of graph-based methods. In our experiments, Graph Neural Networks (GNNs) are used solely for user sentiment classification. The application of GNNs to parody detection and comment sentiment classification remains unexplored, primarily due to the lack of paradigms that allow GNNs to classify edges in graphs. Future work could focus on designing GNN models tailored to edge classification, enabling more comprehensive experiments on parody detection and comment sentiment analysis.
\end{itemize}

\section*{Ethics Statement}
Our proposed benchmark, FanChuan, adheres to the ACL Code of Ethics. All the coauthors also work as annotators, and are compensated at an average hourly rate of 20 SGD. The data we collected is licensed under CC BY 4.0 and is used exclusively for academic purposes. It consists of publicly available website comments and does not contain any sensitive or personal information. To protect user privacy, we filtered out any private data during the data collection and organization process, ensuring that the dataset does not include any user-sensitive content. Additionally, recognizing the potential presence of malicious content in user debates, we have removed harmful comments that violate community ethical standards. Regarding the cultural and topical elements in the datasets, our research remains neutral and free from bias, solely focused on academic exploration. Lastly, AI was used to revise the grammar during the paper writing process.

\clearpage
\bibliography{custom}

\clearpage
\appendix

\section{Dataset Details}\label{apd:dataset_details}

\paragraph{Alibaba-Math} A student from a vocational school achieved remarkable results in the Alibaba Mathematics Competition, despite coming from a school with a less prestigious reputation. Many people supported her, seeing her as a symbol of rising from humble beginnings and a testament to female empowerment. However, some other people questioned her achievements, suggesting that she might have cheated based on snippets from TV interviews. This topic sparked heated discussions on the Chinese internet. To persuade others to believe their claims, some skeptics impersonated her supporters and used exaggerated praise, saying things like, ``\begin{CJK}{UTF8}{gbsn}这位同学有实力！阿里巴巴有眼光！ 请阿里巴巴破格录取进入达摩院，助力阿里科技快速发展 \end{CJK}'' ``\textit{(This student has strength! Alibaba has vision! Please grant her an exceptional admission to DAMO Academy to boost Alibaba’s technological growth  )}'' This is a highly complex topic that encompasses mathematics, education, and gender-related controversies. Annotators working with this dataset must not only be familiar with relevant internet memes but also possess a solid understanding of advanced mathematical concepts.

\paragraph{BridePrice} In some parts of China, there is a tradition of giving a bride price to the bride's family upon marriage. Regarding the demands for exorbitant bride prices, some people believe that the bride price serves as a form of security for the bride, providing her with a greater sense of safety in the marriage. Others argue that the bride price has no inherent relation to marital happiness. This has sparked extensive online debates, and to create an absurd and humorous effect, some opponents of the bride price impersonate the supporters and post comments such as: ``\begin{CJK}{UTF8}{gbsn}是的是的，姐妹们千万别乱嫁人，找不到年入百万的千万别嫁，女孩子五十岁都很值钱！\end{CJK}'' \textit{(Ladies, never marry recklessly. If he doesn't make a million a year, don't marry him. Girls are valuable even at fifty!)}
Gender issues, particularly the topic of bride price, have been a widely debated subject on the Chinese internet for a long time. This dataset requires annotators to be well-versed in these discussions and familiar with the associated memes. 

\paragraph{DrinkWater} A technology video creator recently posted a video titled ``\textit{I Made This to Get Everyone to Drink More Water...}'' sparked controversy. In the video, he introduced a complex ``\textit{Water Drinking Battle}'' system designed to encourage hydration through a reward mechanism. Yet, due to the high design cost and limited effectiveness, some viewers questioned its practicality. Some even ironically pretended to support it, leaving comments like ``\begin{CJK}{UTF8}{gbsn}震古烁今，足以开启第五次技术革命\end{CJK}'' ``\textit{(A groundbreaking innovation capable of launching the fifth technological revolution)}'', to express their dissatisfaction.
This video creator has always been a subject of controversy. While he is well known for his content on science and technology, some critics argue that he lacks fundamental engineering literacy. Annotators working with this dataset should have a basic understanding of scientific and technological concepts. 

\paragraph{CS2} In the Counter Strike 2 (CS2) World Championship finals, G2's newly revamped roster showed impressive strength but once again fell to NAVI, who had already defeated them seven times in a row. This loss sparked heated discussions: someone believes that G2 needs more time to build synergy and has promising potential, while others question whether the roster change truly enhances their chances to win, as they still struggle to overcome their "mental block" against NAVI. Some satirical critics even made eye-catching remarks, such as ``\begin{CJK}{UTF8}{gbsn}传奇捕虾人终结了G2的三日王朝\end{CJK}'' ``\textit{(The legendary shrimp catcher ended G2's three-day dynasty)}'', to express doubts about the effectiveness of G2's roster adjustments. Parody comments in this dataset are particularly difficult to identify for those unfamiliar with the background of CS2, as the comments contain terminology of CS2 game and various aliases of teams and players. Annotators must have a strong understanding of these references to accurately interpret the content.

\paragraph{CampusLife} This dataset was collected from a university forum, covering various discussion topics such as dorm life, campus buses, job hunting, and administration. One particular post sparked a heated debate: a student complained about their roommate bringing their girlfriend to stay overnight in the dorm and sought advice on how to address the situation. The comment section included parodic remarks like ``\textit{Jealous?}'', mocking the situation in a humorous yet disapproving tone. Additionally, during the university's open campus day, a poster appeared in a restroom with the title: ``\textit{Applying to our university? Your tuition funds Palestinian genocide.}'' In response, some users posted parodic comments, such as: ``\textit{Every computer on campus is equipped with an Intel processor, and Intel's R\&D center is in Israel! If you want to avoid supporting genocide, switch to a computer with a Zhaoxin CPU immediately!}''

\paragraph{Tiktok-Trump} In a debate titled ``\textit{Can One Awakened Youth Withstand 20 Trump Supporters?}'', a female Trump supporter lost the debate due to her illogical reasoning and subsequently faced criticism from many netizens who deemed her remarks meaningless. Among the critics, some parodically commented, ``\textit{She did a great job bring up solid points}'', to criticize the Trump supporter's lack of logical reasoning ability.

\paragraph{Reddit-Trump} Trump is a highly controversial figure due to his political stance, ideology, and behavior, sparking widespread debate with both supporters and critics. Some opponents use parody to mimic his tone, such as commenting, ``\textit{He's been tested—more than anyone, by the best doctors in the world. They were amazed, and said they'd never seen scores that high. He'll take another if asked, but they said he doesn’t need to. It’s incredible}'', mocking his rhetorical style and contentious image.
\section{Case Study on LLMs}\label{apd:sec_case_study}
To investigate how well LLMs understand parody, we conduct a case study in which LLMs are asked to provide explanations during prediction. Specifically, we construct the prompt by presenting a comment and its associated topic, then ask the LLMs to determine whether the comment is a parody and to explain their reasoning. After receiving the prediction and explanation from the LLMs, we compare the results with the ground truth label and explanation. The results of the case study for \textit{BridePrice}, \textit{Alibaba-Math}, \textit{DrinkWater}, and \textit{CS2} are presented in Tables \ref{tab:casestudy_brideprice}, \ref{tab:casestudy_alibabamath}, \ref{tab:casestudy_drinkwater}, and \ref{tab:casestudy_CS2}, respectively, using four LLMs: ChatGPT-4o \citep{GPT4}, Qwen 2.5 \citep{Qwen2.5}, DeepSeek-V3 \citep{DeepSeek}, and Claude3.5 \citep{Claude}. The results demonstrate:

(1) LLMs struggle with parody detection. For example, the parody comment in Table \ref{tab:casestudy_brideprice} takes an extreme position opposing the viewpoint that a boyfriend should hand over his salary, yet all the LLMs classify this as a non-parody comment. Additionally, the comment in Table \ref{tab:casestudy_CS2}, which directly expresses a dislike toward the G2 team with analysis, is identified as a parody by 3 of the 4 LLMs.

(2) LLMs frequently provide incorrect explanations when identifying parody comments. Even in the case of \textit{DrinkWater}, shown in Table \ref{tab:casestudy_drinkwater}, where all the LLMs successfully identify the comment as a parody, they fail to generate accurate explanations. The explanations indicate that the LLMs rely mostly on the style and tone of the comment, without a deeper understanding of the implicit meaning.

In conclusion, these results suggest that LLMs struggle to understand parody comments, as they both fail to provide accurate predictions and offer misleading explanations. This highlights the need for further development in LLMs for the task of parody detection.

\begin{table}[h]
    \small
    \centering
    \resizebox{1.0\linewidth}{!}{
    \begin{tabular}{p{1.0\linewidth}}
        \toprule
        \textbf{Question} \\
        \begin{CJK}{UTF8}{gbsn}
        对于“男朋友彩礼给的少工资卡也不愿意上交我很犹豫怎么办？”这一话题，有网友评论：“
        这种男的不要也罢。姐妹要相信自己的价值，好男人很多，一定不要委屈自己。
        一生一次他还不愿意付出所有留退路，你觉得那个退路会是谁？消耗掉的可是你无价的青春啊。”\end{CJK} \\
        \begin{CJK}{UTF8}{gbsn}
        请判断这条评论是否为反串，并给出50字以内的解释。\end{CJK} \\
        Regarding the topic "My boyfriend gives little betrothal money and refuses to hand over his salary card—I'm hesitant, what should I do?", a netizen commented:  
        "This kind of man isn't worth it. Sis, believe in your value—there are many good men, never settle for less."  
        "If he won’t give his all for a once-in-a-lifetime event and leaves himself an out, who do you think that’s for? What’s being wasted is your priceless youth." \\
        Please determine whether this comment is satire and provide an explanation within 50 words.\\
        \textbf{Gronud Truth} \\
        \begin{CJK}{UTF8}{gbsn}
        是。这条评论通过使用“姐妹”这一称呼模仿女性视角，并以极端立场全面否定男性。\end{CJK} \\
        Yes. The comment mimics a female perspective by using the term "sister" and takes an extreme stance by completely negating men. \\
        \midrule
        \textbf{ChatGPT-4o} \ \textcolor{decrease}{\faTimes\ Wrong Answer} \\
        \begin{CJK}{UTF8}{gbsn}
        不是。这条评论表达了女性在婚姻中应得到足够的经济保障和重视，强调自我价值和青春无价，反映了对传统彩礼和工资卡上交的重视，并非反串。\end{CJK} \\
        No. This comment expresses that women should be given adequate financial security and attention in marriage, emphasizes self-worth and the pricelessness of youth, and reflects the importance attached to traditional betrothal gifts and salary card submission. It is not a reverse portrayal. \\
        \midrule
        \textbf{Qwen2.5} \ \textcolor{decrease}{\faTimes\ Wrong Answer} \\
        \begin{CJK}{UTF8}{gbsn}
        不是。这条评论明确支持女性追求平等对待，强调自我价值和寻找更合适伴侣的重要性，没有表现出相反意图或讽刺意味。\end{CJK}  \\
        No. This comment clearly supports women's pursuit of equal treatment, emphasizes the importance of self-worth and finding a more suitable partner, and does not show any contrary intention or irony. \\
        \midrule
        \textbf{DeepSeek-V3} \ \textcolor{decrease}{\faTimes\ Wrong Answer} \\
        \begin{CJK}{UTF8}{gbsn}
        不是。这条评论表达了女性在婚姻中应得到足够的经济保障和重视，强调自我价值和青春无价，反映了对传统彩礼和工资卡上交的重视，并非反串。\end{CJK} \\
        No. This comment expresses that women should be given adequate financial security and attention in marriage, emphasizes self-worth and the pricelessness of youth, and reflects the importance attached to traditional betrothal gifts and salary card submission. It is not a reverse portrayal. \\
        \midrule
        \textbf{Claude3.5} \ \textcolor{decrease}{\faTimes\ Wrong Answer} \\
        \begin{CJK}{UTF8}{gbsn}
        不是。该评论语气、用词和观点都符合特定群体的真实想法，包括"姐妹""价值""青春"等关键词的使用方式自然,不存在夸张或讽刺的痕迹。\end{CJK} \\
        No. The tone, words and opinions of the comment are consistent with the real thoughts of a specific group, including the use of keywords such as "sisters", "values" and "youth" in a natural way without any trace of exaggeration or sarcasm. \\
       \bottomrule
    \end{tabular}
    }
    \caption{A case study of LLMs using an example from \textit{BridePrice}.}
    \label{tab:casestudy_brideprice}
\end{table}

\begin{table}[h]
    \small
    \centering
    \resizebox{1.0\linewidth}{!}
    {
    \begin{tabular}{p{1.0\linewidth}}
        \toprule
        \textbf{Question} \\
        \begin{CJK}{UTF8}{gbsn}
        对于“2024年阿里巴巴数学竞赛预赛中，一位来自中等专业学校的女生取得12名的优秀成绩”这一话题，有网友评论道：“这位同学有实力！阿里巴巴有眼光！ 请阿里巴巴破格录取进入达摩院，助力阿里科技快速发展"\end{CJK} \\
        \begin{CJK}{UTF8}{gbsn}
        请判断这条评论是否为反串，并给出50字以内的解释。\end{CJK} \\
        Regarding the topic "My boyfriend gives little betrothal money and refuses to hand over his salary card—I'm hesitant, what should I do?", a netizen commented:  
        "This kind of man isn't worth it. Sis, believe in your value—there are many good men, never settle for less."  
        "If he won’t give his all for a once-in-a-lifetime event and leaves himself an out, who do you think that’s for? What’s being wasted is your priceless youth." \\
        Please determine whether this comment is satire and provide an explanation within 50 words.\\
        \textbf{Gronud Truth} \\
        \begin{CJK}{UTF8}{gbsn}
        是。这条评论通过对这位同学夸张的赞誉以及对阿里巴巴达摩院破格录取的期望，来引起这位同学反对者的争论。\end{CJK} \\
        Yes. This comment aroused controversy among the student’s opponents by exaggerating the praise for the student and expressing the hope that he would be admitted to Alibaba Damo Academy as an exception.\\
        \midrule
        \textbf{ChatGPT-4o} \ \textcolor{increase}{\faCheck\ Correct Answer, \raisebox{-0.7mm}{\includegraphics[width=0.4cm]{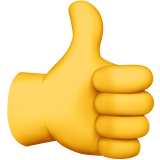}} Good Explanation} \\
        \begin{CJK}{UTF8}{gbsn}
        是，因其夸张用词（如“破格录取”）和不匹配的期待（达摩院门槛极高）。若非反串，也可能是戏谑或调侃，并非严肃建议。\end{CJK} \\
        Yes, due to its exaggerated wording (e.g., "exceptional admission") and unrealistic expectations (since DAMO Academy has extremely high entry requirements). If not parody, it could be playful or mocking rather than a serious suggestion.\\
        \midrule
        \textbf{Qwen2.5} \ \textcolor{decrease}{\faTimes\ Wrong Answer} \\
        \begin{CJK}{UTF8}{gbsn}
        不是。它表达了对姜萍实力的认可和对阿里巴巴的期待，希望阿里能破格录取她，支持科技发展。语气积极正面。\end{CJK}  \\
        No. It expresses recognition of This student’s abilities and expectations for Alibaba, hoping the company will make an exception to admit her and support technological development. The tone is positive and optimistic. \\
        \midrule
        \textbf{DeepSeek-V3} \ \textcolor{increase}{\faCheck\ Correct Answer, \raisebox{-0.7mm}{\includegraphics[width=0.4cm]{images/emoji/+1.png}} Good Explanation} \\
        \begin{CJK}{UTF8}{gbsn}
        是，因其将中专背景与破格进入顶级研究院关联，隐含对选拔标准的不信任，通过夸张建议暗讽现实中的学历门槛或企业宣传策略，点赞表情强化了反讽效果。\end{CJK} \\
        Yes, as it links a vocational school background with exceptional admission to a top research institute, implying skepticism towards the selection criteria. The exaggerated suggestion subtly mocks real-world academic barriers or corporate PR strategies, with the thumbs-up emoji reinforcing the parodic tone. \\
        \midrule
        \textbf{Claude3.5} \ \textcolor{decrease}{\faTimes\ Wrong Answer} \\
        \begin{CJK}{UTF8}{gbsn}
        不是。评论者真诚肯定了姜萍的实力，并对阿里巴巴表达了期望。用语朴实，表达方式也符合普通网友的评论习惯，包括使用表情来强调观点。\end{CJK} \\
        No, The commenter sincerely acknowledges this student's capabilities and expresses expectations for Alibaba. The language is simple and aligns with the typical commenting style of regular netizens, including the use of emojis to emphasize their point. \\
       \bottomrule
    \end{tabular}
    }
    \caption{A case study of LLMs using an example from \textit{Alibaba-Math}.}
    \label{tab:casestudy_alibabamath}
\end{table}

\begin{table}[h]
    \small
    \centering
    \resizebox{1.0\linewidth}{!}
    {
    \begin{tabular}{p{1.0\linewidth}}
        \toprule
        \textbf{Question} \\
        \begin{CJK}{UTF8}{gbsn}
        关于“一位博主为了鼓励员工喝水，将茶水吧改成游戏厅”的话题中，有网友评论：“构思的很好很巧妙，好构思啊，构思.”\end{CJK} \\
        \begin{CJK}{UTF8}{gbsn}
        请判断这条评论是否为反串，并给出50字以内的解释。\end{CJK} \\
        Regarding the topic of "A video creator encouraged employees to drink more water by turning the tea station into a game arcade," a netizen commented: "The idea is well thought out and very clever. Great idea, really great idea."  \\
        Please determine whether this comment is parody and provide an explanation within 50 words. \\
        
        \textbf{Gronud Truth} \\
        \begin{CJK}{UTF8}{gbsn}
        是。这条评论通过"构思"谐音“狗屎”，来嘲讽该博主的方法缺乏实质内容，不切实际。\end{CJK} \\
        Yes. This comment mocks the video creator's method as lacking substance and being impractical by using the Chinese homophone for "idea" which is pronounced as "dog shit". \\
        \midrule
        \textbf{ChatGPT-4o} \ \textcolor{increase}{\faCheck\ Correct Answer}, \raisebox{-0.7mm}{\includegraphics[width=0.4cm]{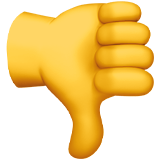}} \textcolor{decrease}{Bad Explanation} \\
        \begin{CJK}{UTF8}{gbsn}
        是。重复“构思”一词带有夸张和调侃意味，结合“很巧妙”可能是在讽刺方案脱离实际，甚至适得其反，体现出一种戏谑或反讽的态度\end{CJK} \\
        Yes. The repetition of the word "idea" carries an exaggerated and mocking tone. Combined with "very clever," it may be satirizing the impracticality of the plan or even suggesting it could backfire, reflecting a playful or ironic attitude.\\
        \midrule
        \textbf{Qwen2.5} \ \textcolor{increase}{\faCheck\ Correct Answer}, \raisebox{-0.7mm}{\includegraphics[width=0.4cm]{images/emoji/_1.png}} \textcolor{decrease}{Bad Explanation} \\
        \begin{CJK}{UTF8}{gbsn}
        是。表面上看似在夸赞，但重复“构思”显得夸张且略带讽刺，暗指将茶水吧改成游戏厅的方案并不合理，实际效果可能适得其反。\end{CJK}  \\
        Yes. While it appears to be a compliment, the exaggerated repetition of "idea" has a slightly ironic tone, implying that turning a tea station into a game arcade is not a reasonable plan and might actually be counterproductive.\\
        \midrule
        \textbf{DeepSeek-V3} \ \textcolor{increase}{\faCheck\ Correct Answer}, \raisebox{-0.7mm}{\includegraphics[width=0.4cm]{images/emoji/_1.png}} \textcolor{decrease}{Bad Explanation} \\
        \begin{CJK}{UTF8}{gbsn}
        是。通过三次重复"构思"制造机械感，配合原事件逻辑悖论（游戏厅促进喝水），构成典型反讽结构——表面夸赞实则暗示想法荒诞，符合中文网络阴阳怪气表达范式。\end{CJK} \\
        Yes The triple repetition of "idea" creates a mechanical effect, combined with the logical paradox of the original event (a game arcade promoting water consumption). This forms a classic structure of irony—appearing as praise but actually hinting at the absurdity of the idea, aligning with the parodic style of Chinese internet humor \\
        \midrule
        \textbf{Claude3.5} \ \textcolor{increase}{\faCheck\ Correct Answer}, \raisebox{-0.7mm}{\includegraphics[width=0.4cm]{images/emoji/_1.png}} \textcolor{decrease}{Bad Explanation} \\
        \begin{CJK}{UTF8}{gbsn}
        是。通过重复使用"构思"一词并且语气刻意夸张，表现出一种模仿式的嘲讽，暗示这个创意可能过于简单或缺乏实质内容。\end{CJK} \\
        Yes. The repeated use of the word "idea" and the deliberately exaggerated tone create a form of imitative mockery, implying that the concept may be overly simplistic or lacking in substance. \\
       \bottomrule
    \end{tabular}
    }
    \caption{A case study of LLMs using an example from \textit{DrinkWater}.}
    \label{tab:casestudy_drinkwater}
\end{table}

\begin{table}[h]
    \small
    \centering
    \resizebox{1.0\linewidth}{!}
    {
    \begin{tabular}{p{1.0\linewidth}}
        \toprule
        \textbf{Question} \\
        \begin{CJK}{UTF8}{gbsn}
        关于“CS2赛事中，G2战队连续7次输给NAVI战队”这一话题，有网友评论：“hunter就是个垃圾，每个战队只能换2人否则取消名额的机制阻碍了90\%战队的重建，也帮他这样的混子混了不少工资。monesy永远不配和zywoo相提并论，抖一步名不虚传，甚至天梯打一个集锦操作就被发到各大平台上，然而这届赛事一张图平均被手撕3次  ”\end{CJK} \\
        \begin{CJK}{UTF8}{gbsn}
        请判断这条评论是否为反串，并给出50字以内的解释。\end{CJK} \\
        In the topic "In CS2 tournaments, G2 has lost to NAVI seven times in a row," a netizen commented:"hunter is just trash. The rule that a team can only replace two players or lose their slot has blocked 90\% of team rebuilds, and has also let freeloaders like him earn quite a bit of salary. monesy will never deserve to be compared to zywoo, best Rifler on TikTok is a well-deserved reputation. He even gets a highlight posted on social media platforms for a single flashy move in ranking, yet in this tournament, he gets destroyed an average of three times per map ."  \\
        Please determine whether this comment is satire and provide an explanation within 50 words.\\
        \textbf{Gronud Truth} \\
        \begin{CJK}{UTF8}{gbsn}
        不是。这条评论通过直接攻击选手（huNter 与 m0NESY）来表达对G2战队和选手的不满与反对。\end{CJK} \\
        No. This comment expresses dissatisfaction and opposition to the G2 team and players by directly attacking the players (hunter and monesy). \\
        \midrule
        \textbf{ChatGPT-4o} \ \textcolor{decrease}{\faTimes\ Wrong Answer} \\
        \begin{CJK}{UTF8}{gbsn}
        是。评论内容极端贬低选手，带有夸张的情绪表达（如“垃圾”“混子”“永远不配”），同时使用表情符号强化讽刺意味，符合常见的反串风格。\end{CJK} \\
        Yes. The content excessively criticizes the players, uses exaggerated emotional expressions (such as "trash" and "freeloader"), and includes emojis to enhance its sarcastic tone, which is characteristic of satirical posts. \\
        \midrule
        \textbf{Qwen2.5} \ \textcolor{decrease}{\faTimes\ Wrong Answer} \\
        \begin{CJK}{UTF8}{gbsn}
        是。评论中使用了夸张的语气和明显的偏见，如“hunter就是个垃圾”和“monesy永远不配和zywoo相提并论”，并且通过讽刺的口吻（如“抖一步名不虚传”）可能意在调侃而非严肃批评，疑似反串黑。\end{CJK}  \\
        Yes, It uses an exaggerated tone and clear bias, such as calling "hunter trash" and claiming "monesy is never worthy of being compared to zywoo." The sarcastic wording (e.g., "shake-and-shoot is well-deserved") suggests an intention to mock rather than genuinely criticize, making it seem like satirical trolling. \\
        \midrule
        \textbf{DeepSeek-V3} \ \textcolor{decrease}{\faTimes\ Wrong Answer} \\
        \begin{CJK}{UTF8}{gbsn}
        是。评论使用夸张贬低言辞（如"垃圾""混子"）、网络热梗（"抖一步"）及表情符号，通过极端负面评价和刻意对比（monesy/zywoo）模仿无脑黑粉语气，意在讽刺而非真实批判。\end{CJK} \\
        Yes. The comment employs exaggerated derogatory terms ("trash," "freeloader"), internet slang ("shake-and-shoot"), and emojis, mimicking the speech style of overly critical fans. The extreme negativity and deliberate comparisons (monesy vs. zywoo) suggest an intent to mock rather than provide a serious critique. \\
        \midrule
        \textbf{Claude3.5} \ \textcolor{increase}{\faCheck\ Correct Answer, \raisebox{-0.7mm}{\includegraphics[width=0.4cm]{images/emoji/+1.png}} Good Explanation} \\
        \begin{CJK}{UTF8}{gbsn}
        不是。解释：用词尖锐且带有明显的情绪色彩，但论点具体且符合CS2玩家的专业表达方式，提到了具体的选手数据和规则限制，显示出对赛事的真实关注和了解。\end{CJK} \\
        No. Explanation: The wording is sharp and emotionally charged, but the arguments are specific and align with the professional discourse of CS2 players. It mentions specific player statistics and rule restrictions, indicating a genuine interest and understanding of the tournament. \\
       \bottomrule
    \end{tabular}
    }
    \caption{A case study of LLMs using an example from \textit{CS2}.}
    \label{tab:casestudy_CS2}
\end{table}
\section{Implementation Details}


In this section, we provide implementation details of all the methods used in Section \ref{sec:Experiments}. Except from Large Language Models (LLMs), all the other methods are trained on 300 epochs, with an early stopping of 5. We use Adam optimizer to update model parameters. The experiments are conducted on a linux server with Ubuntu 20.04, trained on a single NVIDIA RTX A5000 GPU with 24GB memory. All the methods are trained on train set, the hyperparameters are searched on validation set, where the search space is given by:

\begin{itemize}
    \item Hidden Dimension: \{16, 32, 64, 128\},
    \item Learning Rate: \{5e-6, 1e-5, 2e-5, 3e-5, 5e-5, 1e-4\},
    \item Weight Decay: \{1e-5, 1e-4\},
    \item Batch Size: \{16, 32\},
\end{itemize}

For the task of parody detection, the threshold for each dataset is the same for all the methods. Specific, we let the threshold be $0.9415$ for \textit{Alibaba-Math}, $0.9526$ for \textit{BridePrice}, $0.9691$ for \textit{DrinkWater}, $0.9387$ for \textit{CS2}, $0.9262$ for \textit{CampusLife}, $0.9406$ for \textit{Tiktok-Trump}, $0.8768$ for \textit{Reddit-Trump}

Prior to feeding the data into the model, we utilize over sampling with replacement for parody detection, and use Synthetic Minority Over-sampling Technique (SMOTE) \citep{chawla2002smote} for sentiment classification to balance the training data.

Apart from these common settings, we introduce the detailed implementations of each specific model as follows.

\textbf{BoW+MLP} \citep{BoW}
Bag of Words (BoW) is a kind of word embedding method. In this study, the BoW model implemented in Word2Vec \citep{BoW}, aiming to predict a target word based on its surrounding context words. Before using Bag of Words, we standardize text input, remove unnecessary whitespace variations, tokenization text into individual words, and filter out high-frequency words that may not contribute much meaning. Next, we use Bag of Words in Word2Vec to get the word embedding, setting vector size to 50, window to 10, min count to 1, epochs to 50.

Multi-Layer Perceptrons (MLP) is a kind of feedforward neural network. In our study, we employ a three-layer MLP, with a dropout rate set to 0.3 and ReLU as the activation function. 

\textbf{Skip-gram+MLP} \citep{Skip-gram}
Skip-gram is a word embedding method which learns word representations by predicting context words given a target word. Before using Skip-gram, we standardize text input, avoid unnecessary whitespace variations, the text is tokenized into individual words, and filter out high-frequency words that may not contribute much meaning. Then we use Skip-gram in Word2Vec, setting vector size to 50, window to 10, min count to 1, epochs to 50.
The part of MLP is the same as in BoW+MLP.

\textbf{RoBERTa+MLP} \citep{RoBERTa}
RoBERTa ( Robustly Optimized BERT Pretraining Approach ) is an advanced variant of BERT. The part of Next sentence prediction (NSP) is removed from RoBERTa's pre-training objective. To obtain embedding of textual data, we use mean embedding method to compute the average of token embedding from last hidden state. Setting max length to 256, batch size to 32. The part of MLP is the same as in BoW+MLP.

\textbf{BNS-Net} \citep{BNS-Net}
The propagation mechanism in BNS-Net is defined as:$H = f(X, U, W)$, where $X$ represents the textual features, $U$ denotes user embeddings, and $W$ is the weight matrix. The Behavior Conflict Channel (BCC) applies a Conflict Attention Mechanism (CAM) to extract inconsistencies in behavioral patterns, while the Sentence Conflict Channel (SCC) leverages external sentiment knowledge (e.g., SenticNet) to detect implicit and explicit contradictions. BNS-Net is trained using a multi-task loss function, which combines sarcasm classification and sentiment inconsistency modeling:
$L = \lambda_1 J_{\text{sar}} + \lambda_2 J_{\text{imp}} + \lambda_3 J_{\text{exp}} + \lambda_4 J_{\text{balance}}$, where: sar is the sarcasm classification loss,imp and exp correspond to implicit and explicit sentiment contradiction losses. Balance is a balancing term to mitigate bias toward dominant classes. The balancing coefficients used in experiments are: $\lambda_1 = 1.0$, $\quad \lambda_2 = 0.5$, $\quad \lambda_3 = 0.5$, $\quad \lambda_4 = 0.2$.

\textbf{DC-Net} \citep{DC-Net}
The Dual-Channel Network is a dual-channel architecture to realize sarcasm detection by capturing the contrast between literal sentiment and implied sentiment. The model consists of Decomposer, literal channel, implied channel and analyzer. Prior to feeding data into DC-Net, we utilize the opinion lexicon from nltk 3.9.1 to identify the positive and negative word in our datasets. Following the methodology outlined in the original paper, it needs to use GLOVE to obtain the embedding and vocabulary. To generate the literal and implied sentiment labels, we leverage the parody label along with the counts of positive and negative words. These labels are then processed separately in the two channels. Finally the analyzer measure the conflicts between the channels. In our datasets, we follow the original paper and set all of the loss contributions \(\lambda_1\), \(\lambda_2\), \(\lambda_3\) of our DC-Net model are set to 1.

\textbf{QUIET} \citep{QUIET}
The Quantum Sarcasm Model detects sarcasm in text by using quantum-inspired techniques. It converts text and context inputs into dense vector representations through an embedding layer. These embeddings undergo quantum encoding, where sine and cosine functions simulate quantum amplitude and phase encoding, capturing complex relationships. The encoded features are averaged to reduce dimensionality, then passed through a hidden layer with ReLU activation. A sigmoid output layer predicts whether a comment is sarcastic or not. The model addresses class imbalance with class weights and evaluates performance using precision, recall, and F1-score. This single-modality model applies quantum-inspired methods to enhance feature transformation for sarcasm detection.

\textbf{SarcPrompt} \citep{SarcPrompt}
is a prompt-tuning method for sarcasm recognition that enhances PLMs by incorporating prior knowledge of contradictory intentions. The framework comprises two key components: (1) Prompt Construction. (2) Verbalizer Engineering. In our implementation, we adopt the question prompt approach and design bilingual templates tailored to Chinese and English datasets. For Chinese parody detection, we construct the prompt as " \{COMMENT\} \ch{这段话是在反串吗？} \{MASK\}.". For English datasets, we design"\{COMMENT\} Are you parody? \{MASK\}." To enhance model interpretability and alignment with domain knowledge, we employ a verbalizer as paper, where domain-specific label words are mapped based on dataset statistics. In parody detection, we use words like "\ch{反串}", "\ch{是}", "parody", "no". In sentiment classification, we use words like "\ch{支持}", "\ch{反对}", "support", "oppose". The total loss combines cross-entropy (classification) and contrastive losses (enhancing intra-class consistency): $L(\theta) = \lambda_1 L_{\text{sarc}}(\theta) + \lambda_2 L_{\text{con}}(\theta)$, where \(\lambda_1 = 1\) and \(\lambda_2\) is selected from \{0.05, 0.1, 0.2, 0.5, 1\} via validation, following the original paper's hyperparameter selection.

\textbf{GCN} \citep{GCN}
All Graph Neural Networks (GNNs), including GCN, GAT, and GraphSAGE, are implemented using PyTorch Geometric \citep{pyg}, with the version specified as 2.6.1. For the GCN, we set the number of graph convolution layers to 2, the size of the hidden embedding to 64, and the dropout rate to 0.5. Additionally, we incorporate residual connections \citep{residual} and layer normalization \citep{layer_norm} to enhance model performance, as suggested by \citet{classic_gnn_strong}.

\textbf{GAT} \citep{GAT}
In GAT, we adopt the same configuration as in Graph Convolutional Networks (GCN), utilizing 2 graph convolution layers, a hidden embedding size of 64, and a dropout rate of 0.5. Additionally, we set the number of attention heads to 8.

\textbf{GraphSAGE} \citep{GraphSAGE}
In GraphSAGE, we adopt the same configuration as in Graph Convolutional Networks (GCN), utilizing 2 graph convolution layers, a hidden embedding size of 64, and a dropout rate of 0.5. Additionally, we set the neighborhood size to 5.

\textbf{LLMs}
we employ a variety of LLMs from different companies to perform parody detection and sentiment classification, which include ChatGPT-4o (and 4o-mini) \citep{GPT4}, ChatGPT-o1-mini \citep{ChatGPT-o1}, ChatGPT-o3-mini \citep{ChatGPT-o3} Claude 3.5 \citep{Claude}, Qwen 2.5 \citep{Qwen2.5}, DeepSeek-V3 \citep{DeepSeek}, and DeepSeek-R1 \citep{DeepSeek-R1}.They require different kinds of input formats, objects and parameters. Except reasoning model, we set temperature to 0, which reasoning model not support this object. For reasoning model, they have to use more and more tokens to complete the reasoning procedure before outputting the content. To optimize model performance, we design task-specific prompts, ensuring that each LLM receives input formulations tailored to the characteristics of parody detection and sentiment analysis. For example, in parody detection, we design the prompt as \textit{``You are a helpful assistant trained to classify whether a statement is parody or not.''} in the system role, and \textit{``Determine whether the following comment is parody:\{text\}\texttt{\textbackslash n} Directly output 1 for parody, 0 for non-parody.''} in the user role. In particular, ChatGPT o1-mini doesn't have the system role, so we input all in the user role.



%
\section{Additional Results}

This section introduces additional results in our experiments. We introduce more results of the influence of context to parody detection in Section \ref{apd:impact_context} and the influence of parody to sentiment classification in Section \ref{apd:impact_parody}. Then, we show the performance comparison of reasoning LLMs and non-reasoning LLMs in Section \ref{apd:reasoning_llms}. Last, we investigate the impact of train ratio of embedding-based models compared with LLMs in Section \ref{apd:train_ratio}.

\subsection{Influence of Context to Parody Detection}\label{apd:impact_context}
\begin{figure}[htbp]
  \centering
  \begin{subfigure}[b]{0.23\textwidth}
    \centering
    \includegraphics[width=\textwidth]{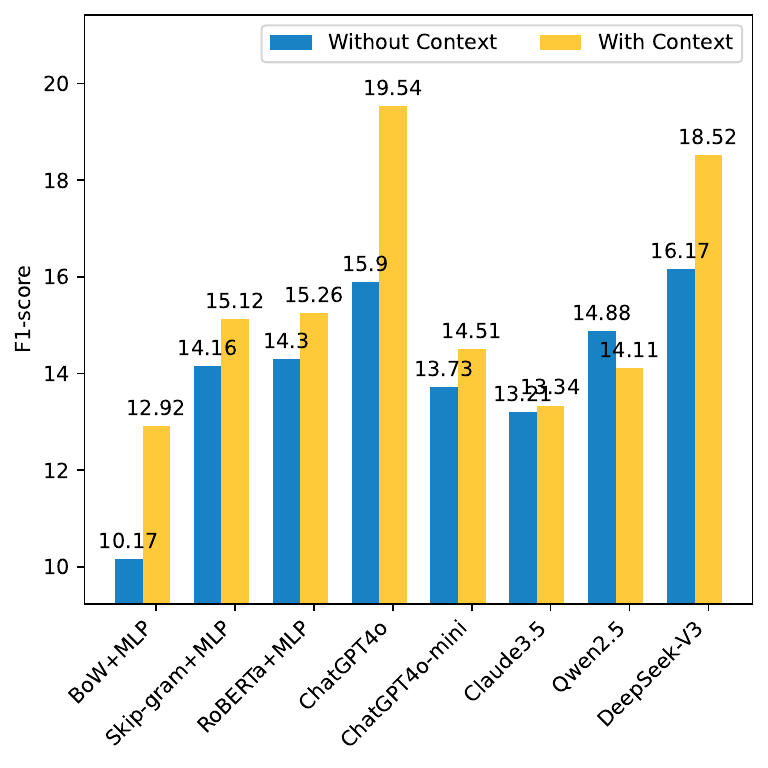}
    \caption{Alibaba-Math}
    \label{fig:compare_context_sub1}
  \end{subfigure}
  \hfill
  \begin{subfigure}[b]{0.23\textwidth}
    \centering
    \includegraphics[width=\textwidth]{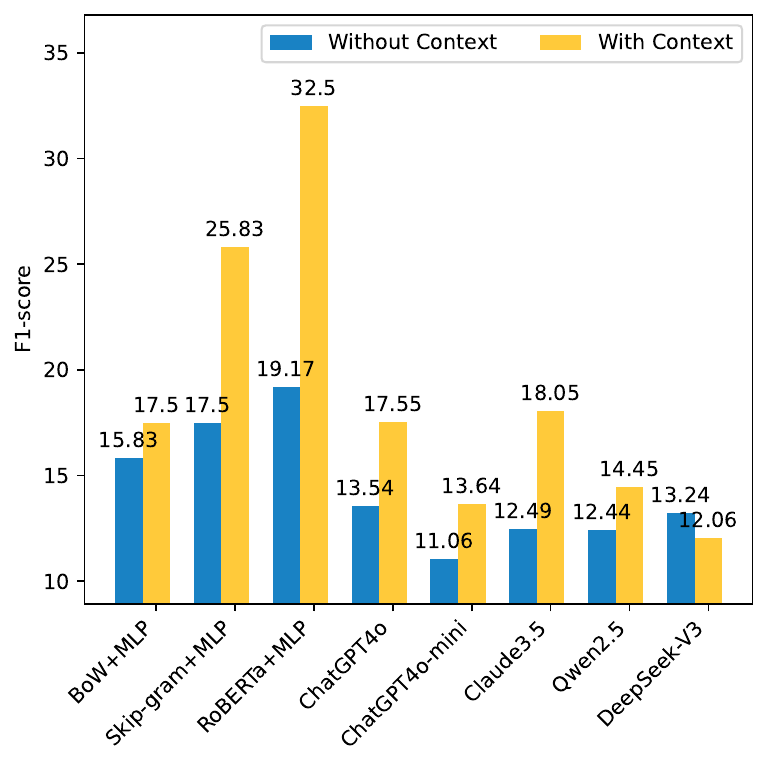}
    \caption{BridePrice}
    \label{fig:compare_context_sub2}
  \end{subfigure}
  
  \vspace{0.7cm}  
  
  \begin{subfigure}[b]{0.23\textwidth}
    \centering
    \includegraphics[width=\textwidth]{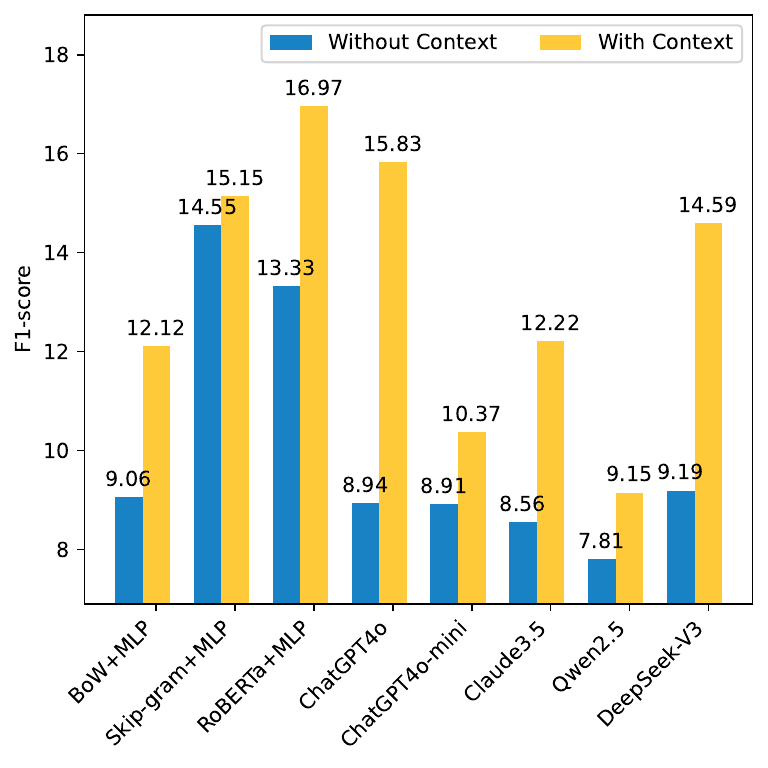}
    \caption{DrinkWater}
    \label{fig:compare_context_sub3}
  \end{subfigure}
  \hfill
  \begin{subfigure}[b]{0.23\textwidth}
    \centering
    \includegraphics[width=\textwidth]{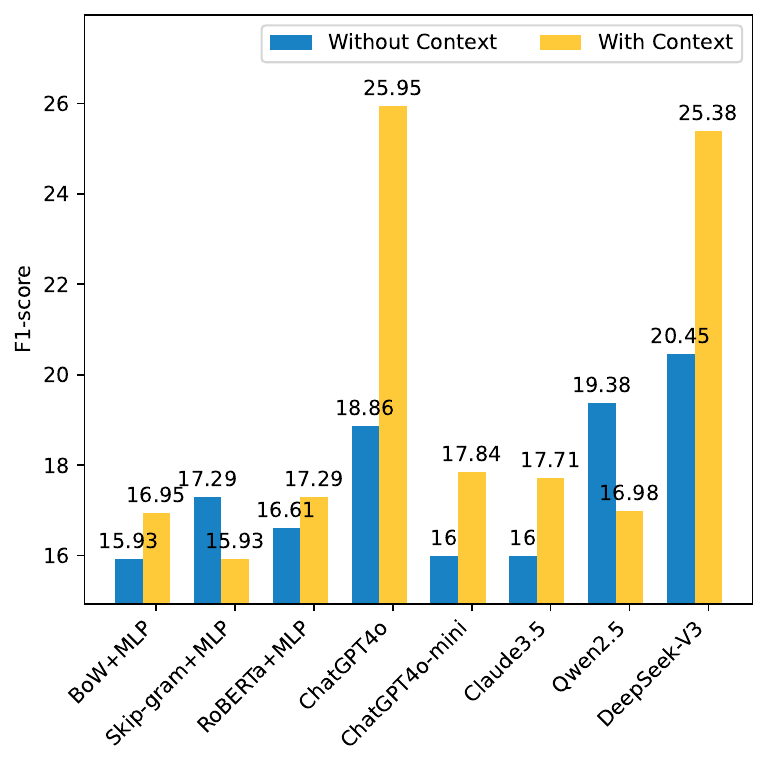}
    \caption{CS2}
    \label{fig:compare_context_sub4}
  \end{subfigure}

  \vspace{0.7cm}  

  \begin{subfigure}[b]{0.23\textwidth}
    \centering
    \includegraphics[width=\textwidth]{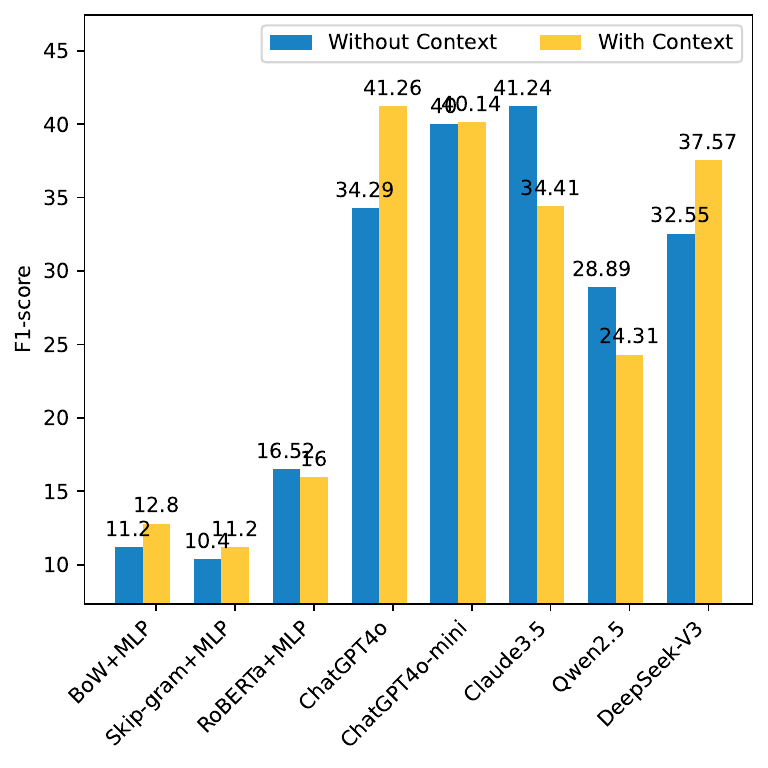}
    \caption{CampusLife}
    \label{fig:compare_context_sub5}
  \end{subfigure}
  \hfill
  \begin{subfigure}[b]{0.23\textwidth}
    \centering
    \includegraphics[width=\textwidth]{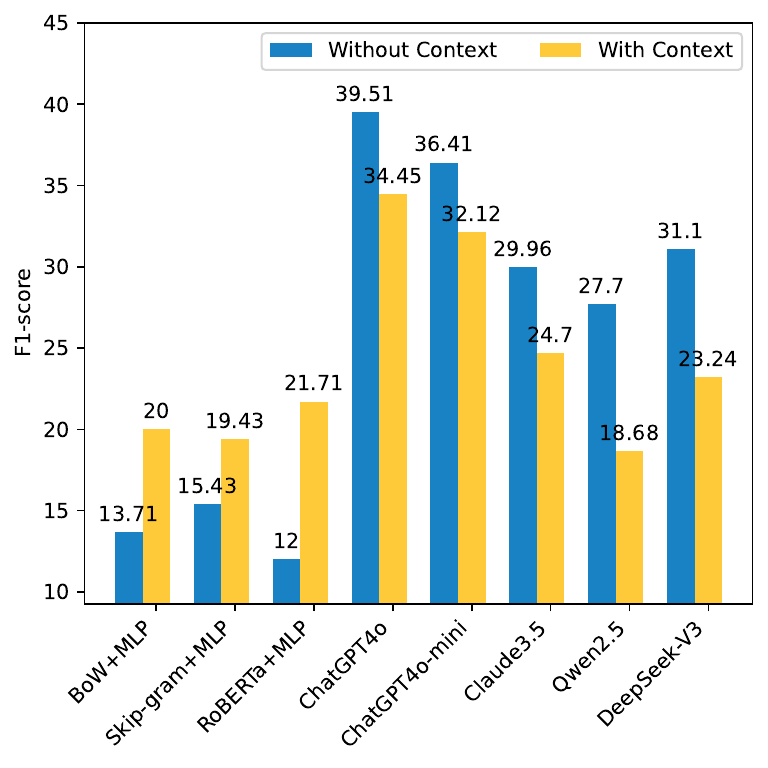}
    \caption{Tiktok-Trump}
    \label{fig:compare_context_sub6}
  \end{subfigure}

  \vspace{0.7cm}  

  \begin{subfigure}[b]{0.23\textwidth}
    \centering
    \includegraphics[width=\textwidth]{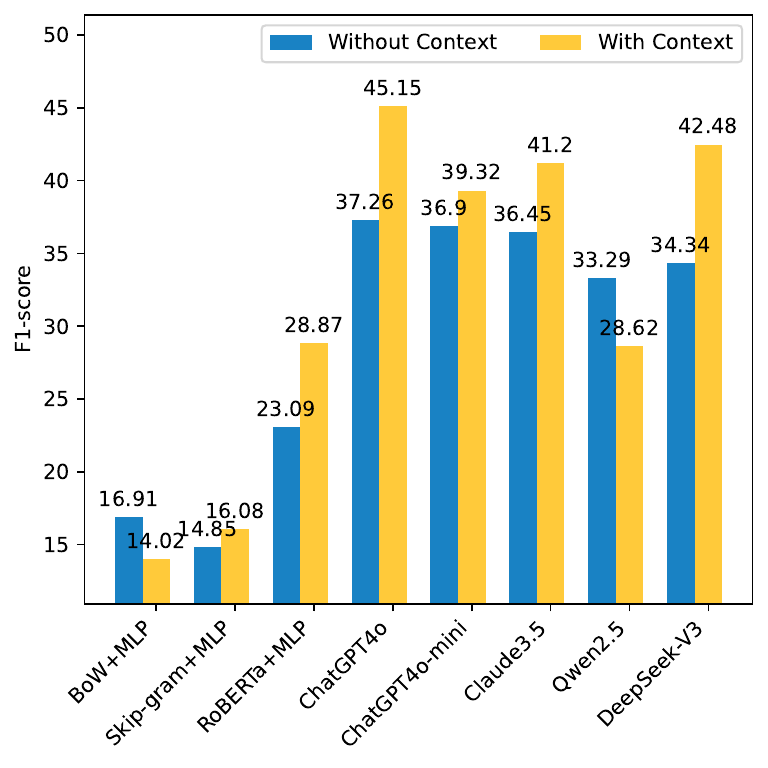}
    \caption{Reddit-Trump}
    \label{fig:compare_context_sub7}
  \end{subfigure}

  \vspace{0.7cm}  

  \caption{Impact of contextual information on parody detection across seven datasets.}
  \label{fig:apd_impact_context}
\end{figure}

Figure \ref{fig:apd_impact_context} illustrates the detailed results of the performance comparison of the F1 score in parody detection with and without context across seven datasets. Generally, contextual information significantly enhances model performance on most datasets and methods. For instance, on \textit{Alibaba-Math}, the performance of ChatGPT4o improves from $15.9$ to $19.54$, while on \textit{BridePrice}, the performance of RoBERTa+MLP increases from $19.17$ to $32.50$. These results indicate that contextual information is beneficial for parody detection. This finding aligns with the results in \citet{dialogue_bamman, dialogue_wang}, which show that providing dialogue as context significantly improves model performance in sarcasm detection.

However, although contextual information significantly improves model performance on most datasets, there are still some datasets where context does not enhance or even decreases model performance. For example, on \textit{Tiktok-Trump}, the model performance decreases, and on \textit{CampusLife}, the performance remains similar after adding contextual information. This suggests that contextual information may not always contribute to improving model performance in parody detection.

\subsection{Influence of Parody to Sentiment Classification}\label{apd:impact_parody}
\begin{figure}[htbp]
  \centering
  \begin{subfigure}[b]{0.23\textwidth}
    \centering
    \includegraphics[width=\textwidth]{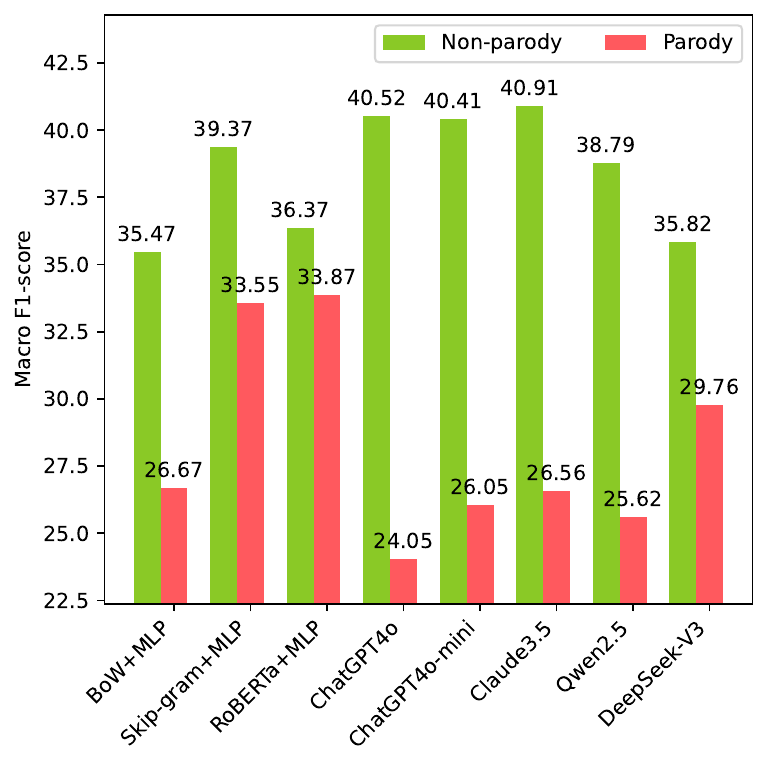}
    \caption{Alibaba-Math}
    \label{fig:ORPtoSenti_sub1}
  \end{subfigure}
  \begin{subfigure}[b]{0.23\textwidth}
    \centering
    \includegraphics[width=\textwidth]{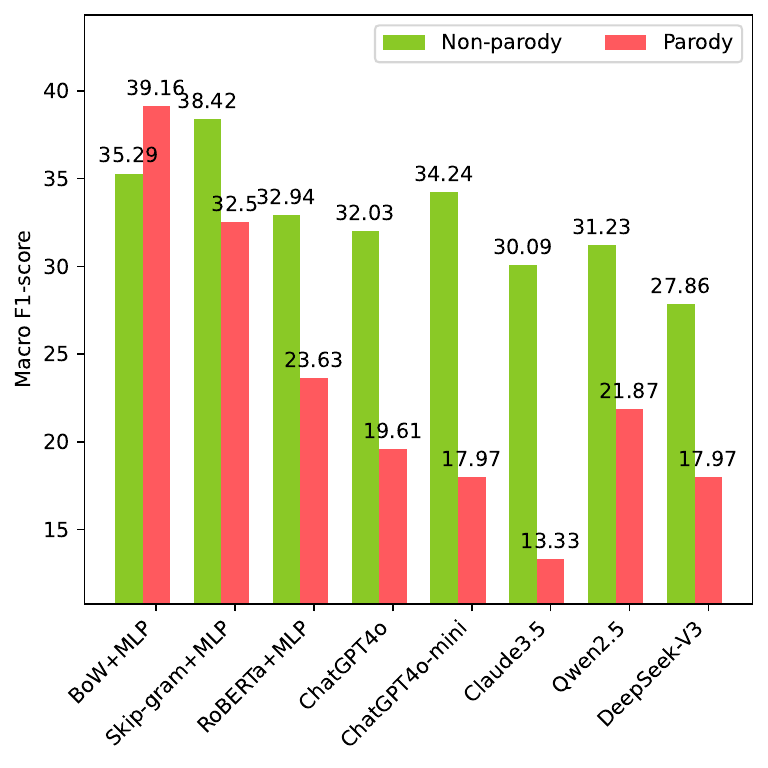}
    \caption{BridePrice}
    \label{fig:ORPtoSenti_sub2}
  \end{subfigure}

  \vspace{0.7cm}
  
  \begin{subfigure}[b]{0.23\textwidth}
    \centering
    \includegraphics[width=\textwidth]{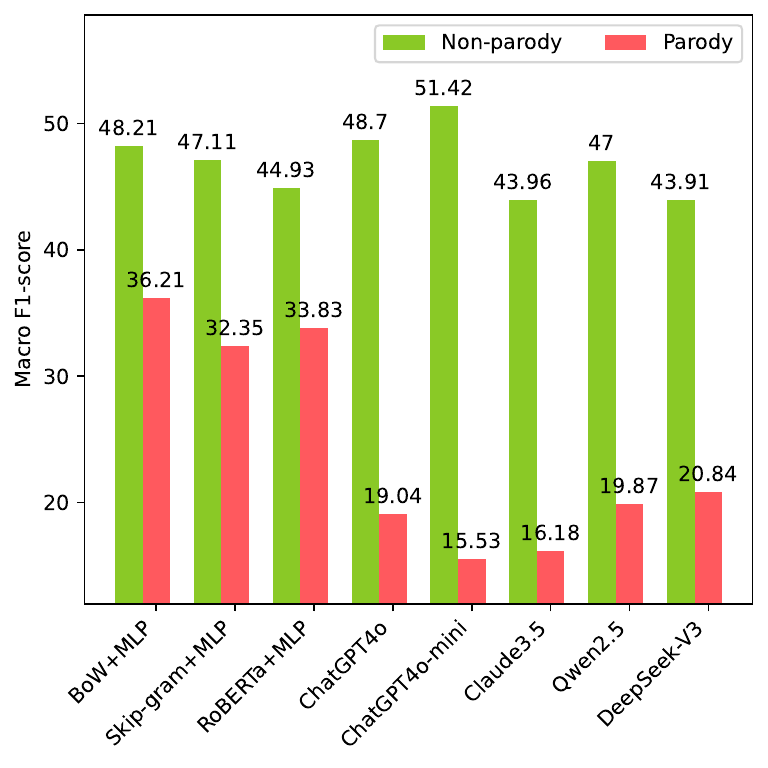}
    \caption{DrinkWater}
    \label{fig:ORPtoSenti_sub3}
  \end{subfigure} 
  \begin{subfigure}[b]{0.23\textwidth}
    \centering
    \includegraphics[width=\textwidth]{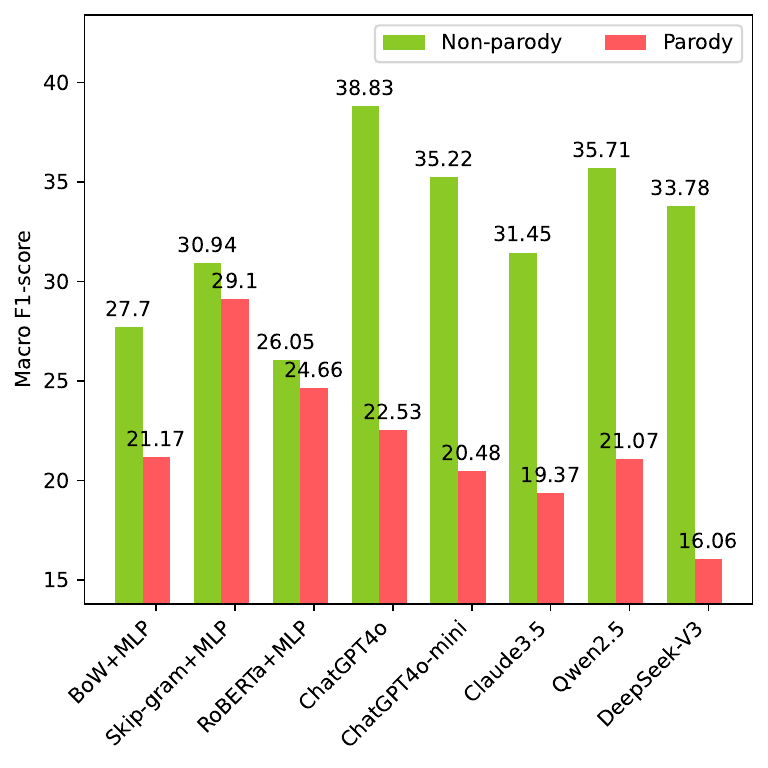}
    \caption{CS2}
    \label{fig:ORPtoSenti_sub4}
  \end{subfigure}

  \vspace{0.7cm}

  \begin{subfigure}[b]{0.23\textwidth}
    \centering
    \includegraphics[width=\textwidth]{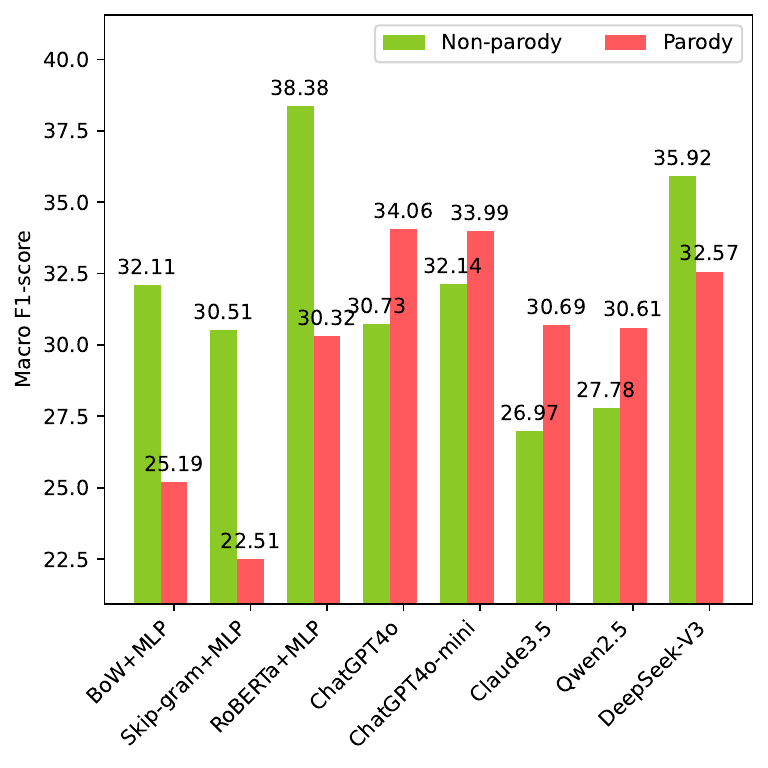}
    \caption{CampusLife}
    \label{fig:ORPtoSenti_sub5}
  \end{subfigure}
  \begin{subfigure}[b]{0.23\textwidth}
    \centering
    \includegraphics[width=\textwidth]{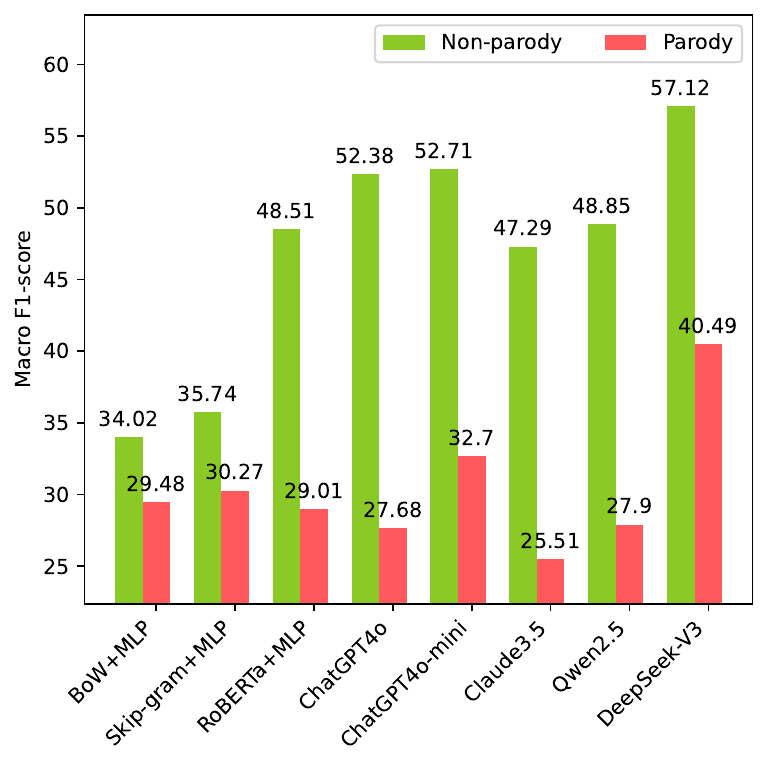}
    \caption{Tiktok-Trump}
    \label{fig:ORPtoSentit_sub6}
  \end{subfigure}

  \vspace{0.7cm}

  \begin{subfigure}[b]{0.23\textwidth}
    \centering
    \includegraphics[width=\textwidth]{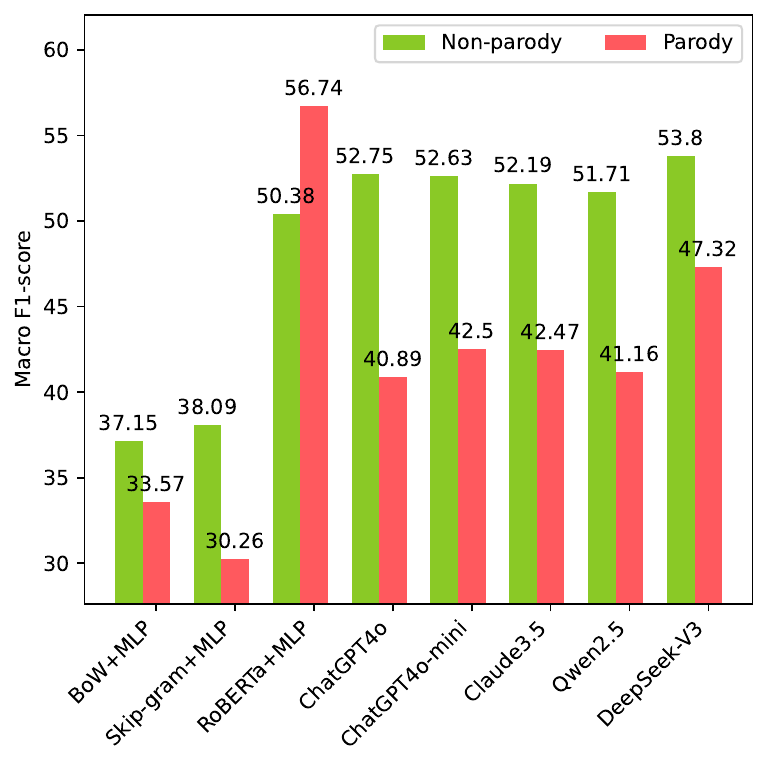}
    \caption{Reddit-Trump}
    \label{fig:ORPtoSenti_sub7}
  \end{subfigure}

  \vspace{0.7cm}  

  \caption{Impact of parody on comment sentiment classification across seven datasets.}
  \label{fig:apd_impact_parody}
\end{figure}

Figure \ref{fig:apd_impact_parody} presents the detailed model performance of comment sentiment classification on parody and non-parody comments across seven datasets. In the \textit{DrinkWater} dataset, large language models (LLMs) such as ChatGPT-4o-mini (F1-score: 51.42) and Qwen2.5 (F1-score: 47.00) achieve competitive performance compared to embedding-based methods like Bag of Words (BoW) (F1-score: 48.21), Skip-gram (F1-score: 47.11), and RoBERTa (F1-score: 44.93) when parody is not present. However, for parody comments, the performance of LLMs degrades significantly, falling below that of embedding-based approaches. For instance, ChatGPT-4o drops from an F1-score of 48.7 to 19.04, and ChatGPT-4o-mini declines from 51.42 to 15.53, whereas embedding-based methods exhibit greater robustness, with BoW decreasing from 48.21 to 36.21, Skip-gram from 47.11 to 32.35, and RoBERTa from 44.93 to 33.83. Overall, these results indicate that parody presents substantial challenges for sentiment classification, and LLMs struggle to maintain their advantage over traditional embedding-based methods in this context.

\subsection{Reasoning LLMs in Parody Detection}\label{apd:reasoning_llms}
\begin{figure}[htbp]
  \centering
  \begin{subfigure}[b]{0.23\textwidth}
    \centering
    \includegraphics[width=\textwidth]{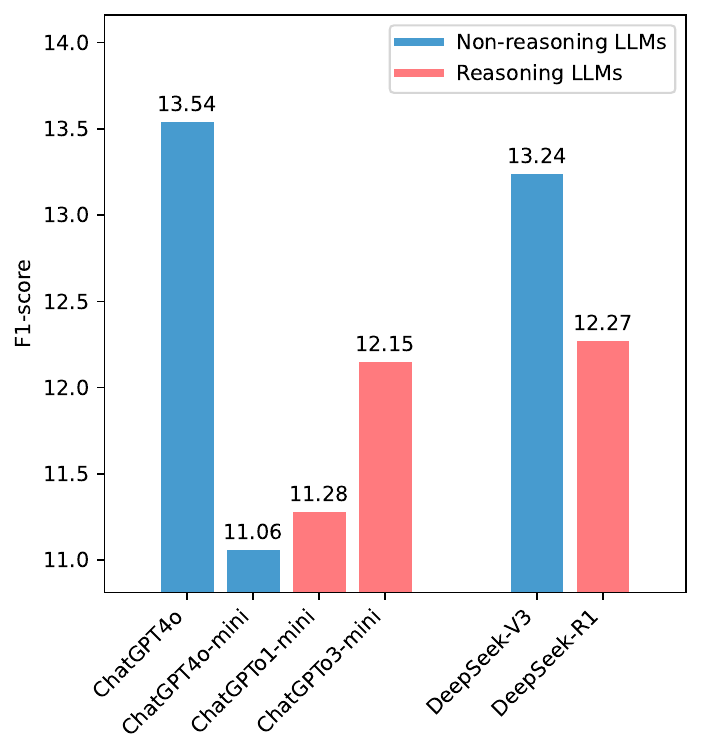}
    \caption{BridePrice}
    \label{fig:LLM_compare_sub2}
  \end{subfigure}
  \hfill
  \begin{subfigure}[b]{0.23\textwidth}
    \centering
    \includegraphics[width=\textwidth]{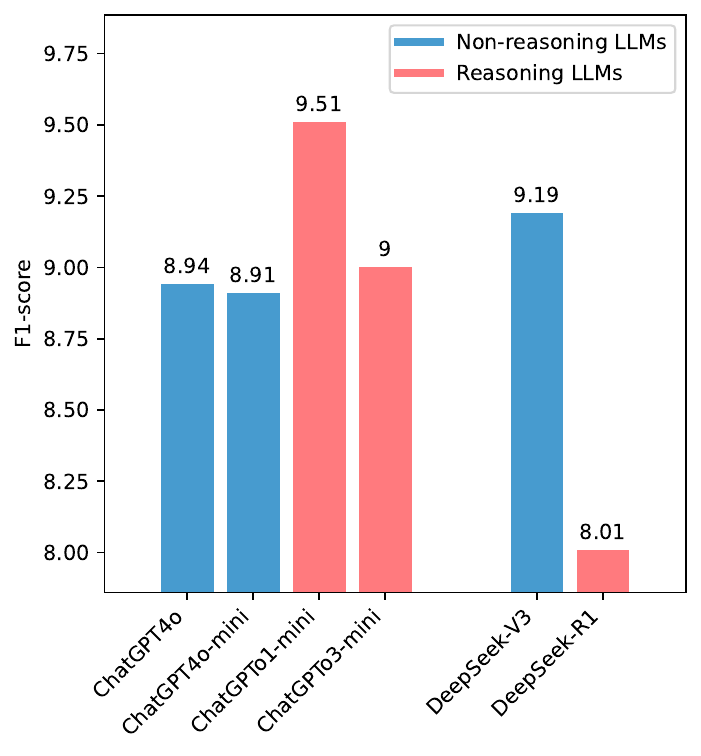}
    \caption{DrinkWater}
    \label{fig:LLM_compare_sub3}
  \end{subfigure}

  \vspace{0.7cm}  

  \begin{subfigure}[b]{0.23\textwidth}
    \centering
    \includegraphics[width=\textwidth]{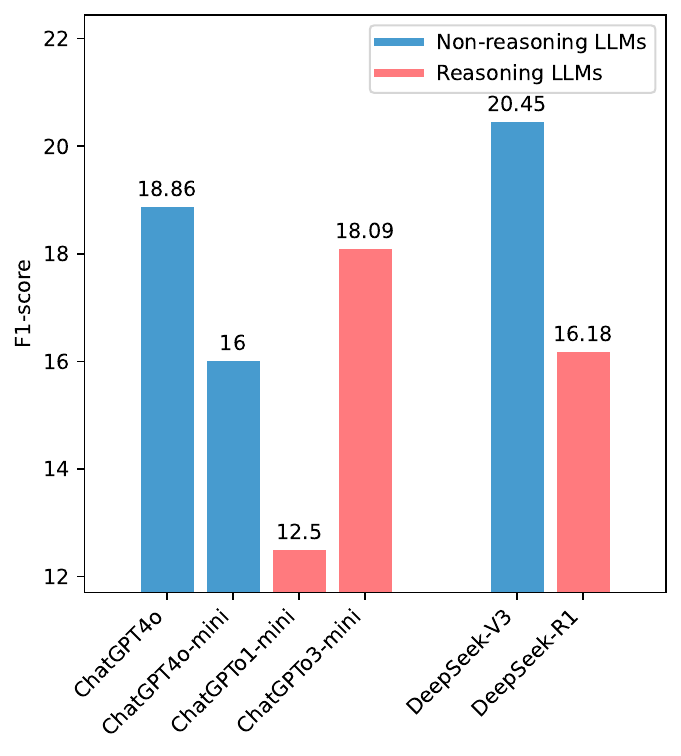}
    \caption{CS2}
    \label{fig:LLM_compare_sub4}
  \end{subfigure}
  \hfill
  \begin{subfigure}[b]{0.23\textwidth}
    \centering
    \includegraphics[width=\textwidth]{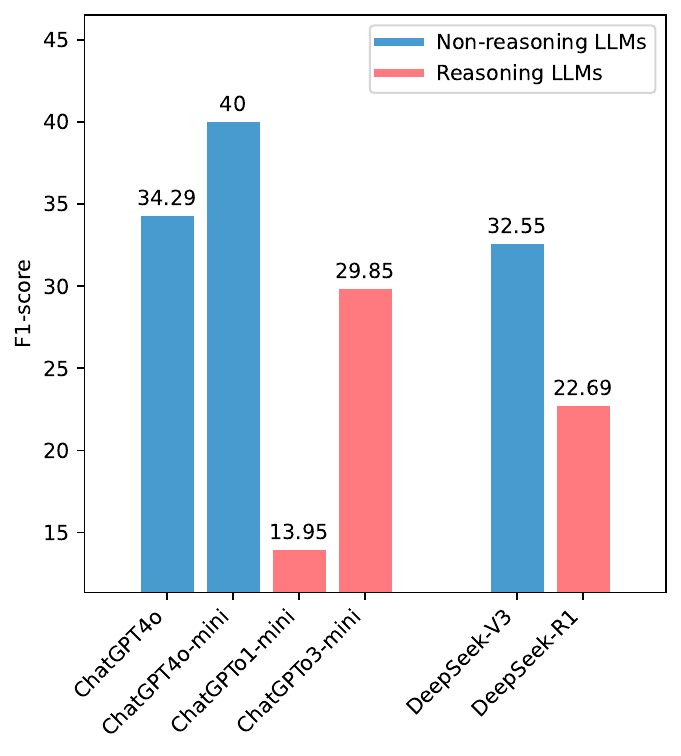}
    \caption{CampusLife}
    \label{fig:LLM_compare_sub5}
  \end{subfigure}

  \vspace{0.7cm}  

  \begin{subfigure}[b]{0.23\textwidth}
    \centering
    \includegraphics[width=\textwidth]{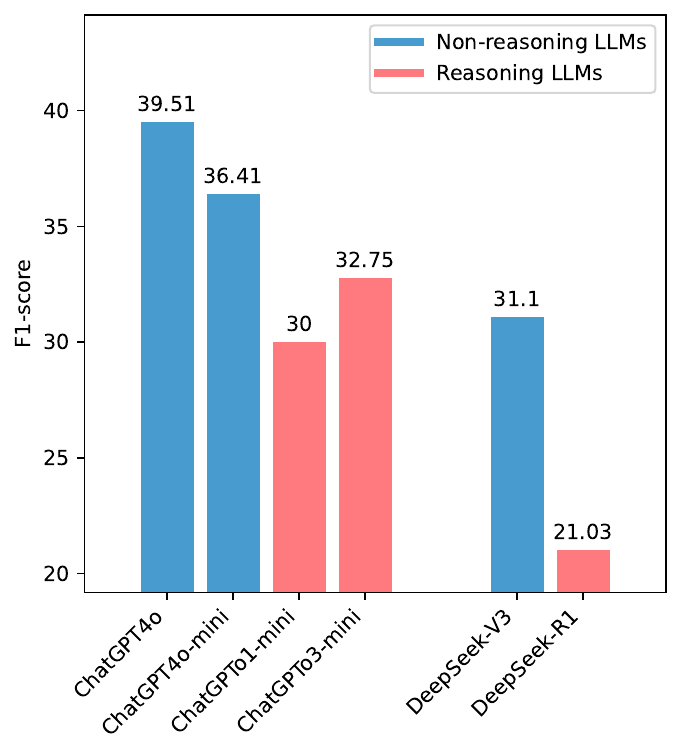}
    \caption{Tiktok-Trump}
    \label{fig:LLM_compare_sub6}
  \end{subfigure}
  \hfill
  \begin{subfigure}[b]{0.23\textwidth}
    \centering
    \includegraphics[width=\textwidth]{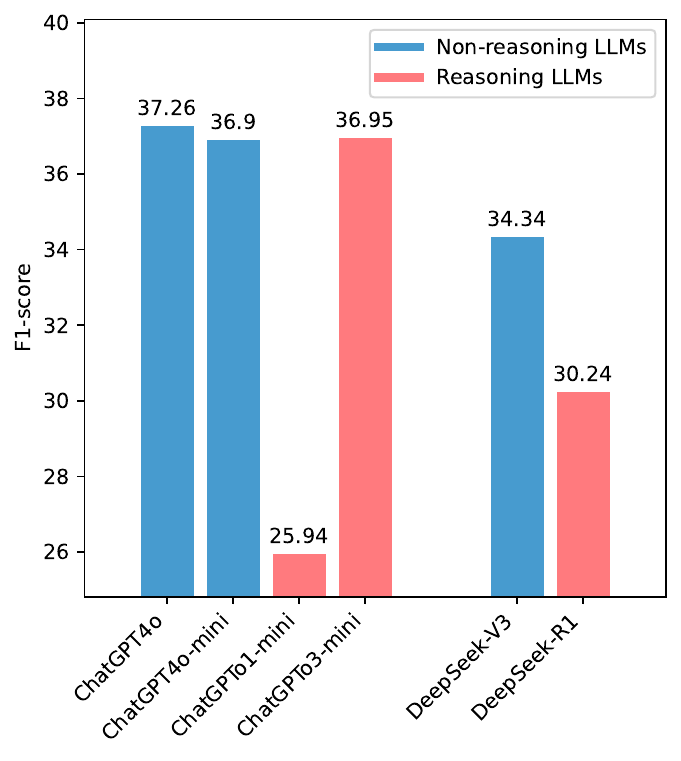}
    \caption{Reddit-Trump}
    \label{fig:LLM_compare_sub7}
  \end{subfigure}

  \vspace{0.7cm}  

  \caption{A Comparative Performance Analysis of Reasoning vs. Non-Reasoning LLMs}
  \label{fig:apd_reasoning_llms}
\end{figure}

We present the details of reasoning LLMs in parody detection across six datasets in Figure \ref{fig:apd_reasoning_llms}. Our findings indicate that reasoning LLMs do not exhibit a performance advantage compared to non-reasoning LLMs. For instance, ChatGPT-o1-mini and ChatGPT-o3-mini underperform relative to ChatGPT4o-mini on the \textit{CampusLife} and \textit{Tiktok-Trump} datasets. Additionally, DeepSeek-R1 significantly underperforms compared to DeepSeek-V3 across all datasets. 

These results suggest that reasoning does not enhance LLM performance in parody detection. We speculate that this may be due to the nature of parody, which often relies on indirect or subtle cues related to tone, context, and nuance rather than direct logical inference. In such cases, non-reasoning LLMs, which excel at identifying statistical patterns and linguistic structures, may be more effective at detecting parody than reasoning LLMs that focus excessively on logical steps or detailed analysis.

\subsection{Impact of Supervision Ratio}\label{apd:train_ratio}

\begin{figure}[h]
  \begin{subfigure}[b]{0.23\textwidth}
    \centering
    \includegraphics[width=\textwidth]{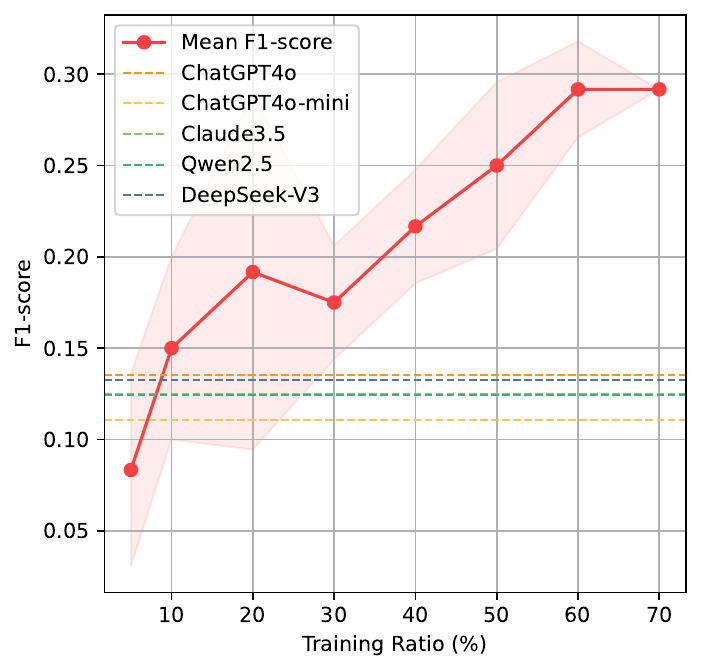}
    \caption{BridePrice}
    \label{fig:parody_sub2}
  \end{subfigure}
  \hfill
  \begin{subfigure}[b]{0.23\textwidth}
    \centering
    \includegraphics[width=\textwidth]{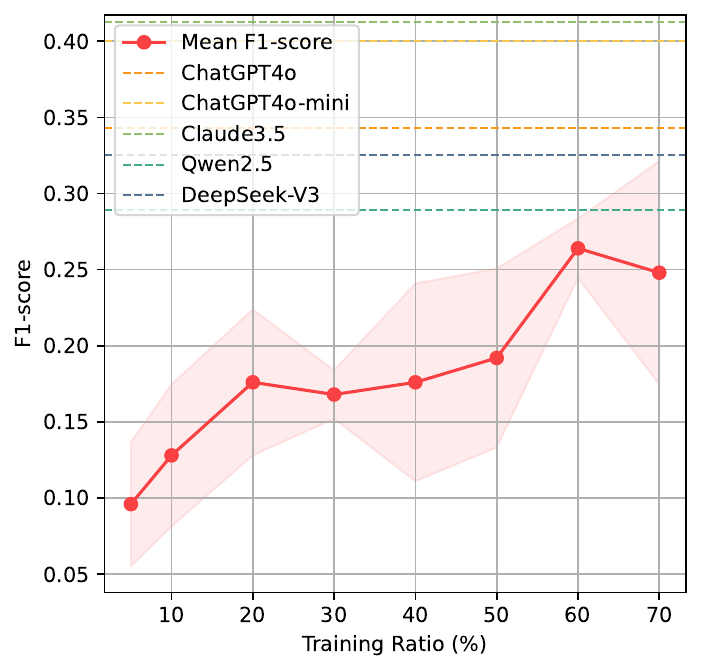}
    \caption{CampusLife}
    \label{fig:parody_sub5}
  \end{subfigure}
  
  \vspace{0.7cm}

  \caption{Impact of training ratio to RoBERTa+MLP on parody detection}
  \label{fig:train_ratio_parody_detection}
\end{figure}

\begin{figure}[h]
  \centering
  \begin{subfigure}[b]{0.23\textwidth}
    \centering
    \includegraphics[width=\textwidth]{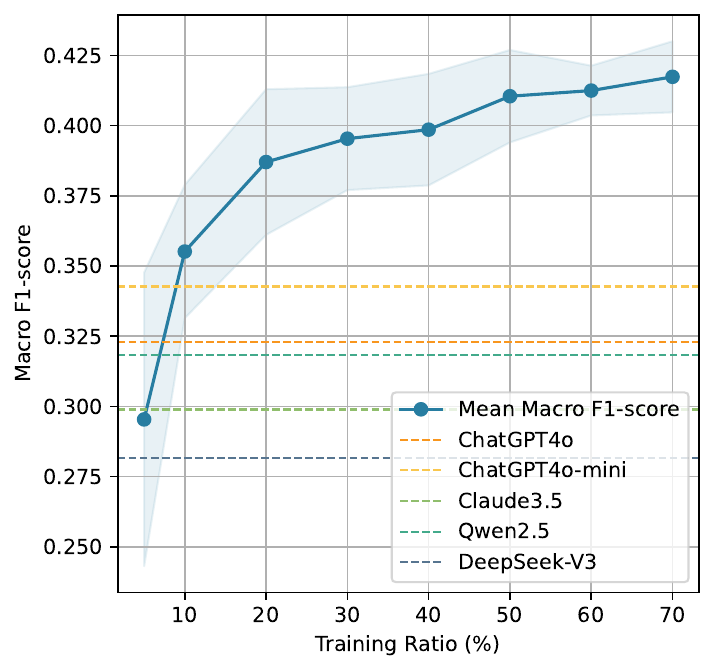}
    \caption{BridePrice}
    \label{fig:sentiment_sub2}
  \end{subfigure}
  \hfill
  \begin{subfigure}[b]{0.23\textwidth}
    \centering
    \includegraphics[width=\textwidth]{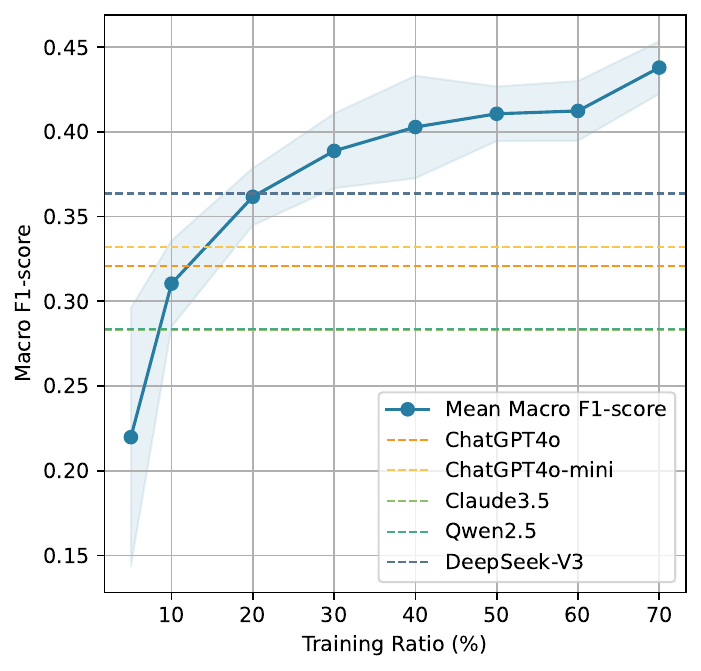}
    \caption{CampusLife}
    \label{fig:sentiment_sub5}
  \end{subfigure}

  \vspace{0.7cm}  

  \caption{Impact of training ratio to RoBERTa+MLP on comment sentiment classification}
  \label{fig:train_ratio_comment_senti}
\end{figure}
\begin{figure}[h]
  \centering
  \begin{subfigure}[b]{0.23\textwidth}
    \centering
    \includegraphics[width=\textwidth]{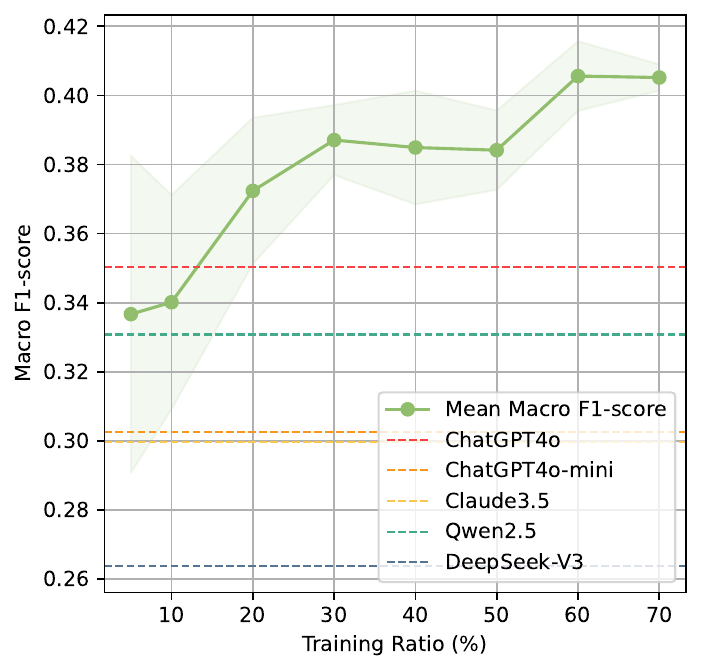}
    \caption{BridePrice}
    \label{fig:user_sentiment_sub2}
  \end{subfigure}
  \hfill
  \begin{subfigure}[b]{0.23\textwidth}
    \centering
    \includegraphics[width=\textwidth]{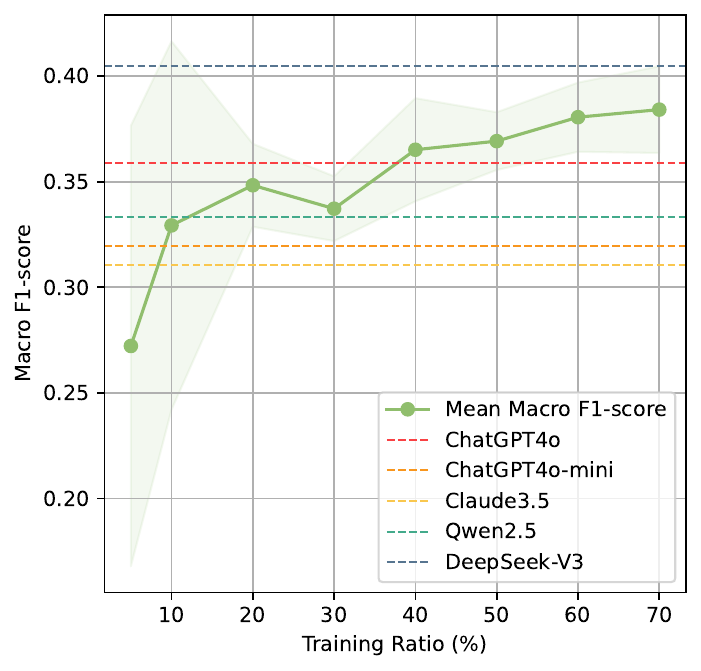}
    \caption{CampusLife}
    \label{fig:user_sentiment_sub5}
  \end{subfigure}

  \vspace{0.7cm}  
  
  \caption{Impact of training ratio to RoBERTa+MLP on user sentiment classification}
  \label{fig:train_ratio_user_senti}
\end{figure}

The embedding-based methods used in our experiments require explicit training on labeled data, whereas LLMs like RoBERTa do not require such training once pre-trained. Therefore, the performance of embedding-based models depends on the size and quality of the training set. To explore this, we investigate how varying the training ratio influences model performance by gradually increasing the training set size while keeping the test set constant. The results for RoBERTa+MLP under different train ratio are presented in Figures \ref{fig:train_ratio_parody_detection}, \ref{fig:train_ratio_comment_senti}, and \ref{fig:train_ratio_user_senti} for parody detection, comment sentiment classification, and user sentiment classification. In all tasks, we observe that the performance increases monotonically with the training ratio, highlighting the benefit of additional training data for embedding-based methods.

In addition, on the \textit{BridePrice} dataset, only $10\%$ supervision is enough for RoBERTa to outperform all LLMs in parody detection, indicating a limitation of LLMs in domain-specific tasks. This suggests that fine-tuned models like RoBERTa perform better with minimal supervision in specialized contexts. In contrast, on the \textit{CampusLife} dataset, RoBERTa's performance consistently falls below that of all LLMs, regardless of the training ratio. This suggests that LLMs are more effective in tasks requiring generalizable knowledge and flexibility, such as parody detection in diverse, context-rich domains. These results demonstrate that LLMs remain powerful in specific areas requiring flexibility in adapting to diverse linguistic contexts and nuanced understanding, while embedding-based models like RoBERTa excel in more targeted, domain-specific tasks.


\end{document}